\documentclass[10pt,twocolumn,letterpaper]{article}

\usepackage{iccv}

\usepackage{times}
\usepackage{epsfig}
\usepackage{graphicx}
\usepackage{amsmath}
\usepackage{amssymb}
\usepackage{bm}
\usepackage{subcaption}
\usepackage{booktabs}
\usepackage{multirow}
\usepackage{tabularx}
\usepackage{makecell}
\usepackage[accsupp]{axessibility}

\usepackage{amsthm}

\usepackage{xcolor}

\makeatletter\renewcommand\paragraph{\@startsection{paragraph}{4}{\z@}
  {.5em \@plus1ex \@minus.2ex}{-.5em}{\normalfont\normalsize\bfseries}}\makeatother

\usepackage[pagebackref=true,breaklinks=true,letterpaper=true,colorlinks,bookmarks=false]{hyperref}

\iccvfinalcopy %

\def\iccvPaperID{2623} %

\begin{document}
	
\title{High-Resolution Optical Flow from 1D Attention and Correlation}

\author{{Haofei Xu\textsuperscript{1}\thanks{Work primarily done while interning at MSRA mentored by JY} \quad Jiaolong Yang\textsuperscript{2} \quad Jianfei Cai\textsuperscript{3}  \quad Juyong Zhang\textsuperscript{1}\quad Xin Tong\textsuperscript{2}} \\

{\normalsize \textsuperscript{1}University of Science and Technology of China \quad \textsuperscript{2}Microsoft Research Asia} 
\\ 
{\normalsize \textsuperscript{3}Department of Data Science and AI, Monash University}
\\ 
{\tt\footnotesize \{xhf@mail., juyong@\}ustc.edu.cn \hspace{1 mm} \{jiaoyan, xtong\}@microsoft.com \hspace{1 mm} jianfei.cai@monash.edu}

}

\maketitle

\begin{abstract}
	Optical flow is inherently a 2D search problem, and thus the computational complexity grows quadratically with respect to the search window, making large displacements matching infeasible for high-resolution images. In this paper, we take inspiration from Transformers and propose a new method for high-resolution optical flow estimation with significantly less computation. Specifically, a 1D attention operation is first applied in the vertical direction of the target image, and then a simple 1D correlation in the horizontal direction of the attended image is able to achieve 2D correspondence modeling effect. The directions of attention and correlation can also be exchanged, resulting in two 3D cost volumes that are concatenated for optical flow estimation. The novel 1D formulation empowers our method to scale to very high-resolution input images while maintaining competitive performance. Extensive experiments on Sintel, KITTI and real-world 4K ($2160 \times 3840$) resolution images demonstrated the effectiveness and superiority of our proposed method. Code and models are available at \url{https://github.com/haofeixu/flow1d}.
	
\end{abstract}

\vspace{-8pt}

\section{Introduction}

Optical flow estimation, a classic topic in computer vision, is a fundamental building block of various real-world applications such as 3D reconstruction \cite{li2019learning}, video processing \cite{jiang2018super} and action recognition \cite{simonyan2014two}. The recent advancement of deep learning enables directly optical flow learning with a neural network \cite{ilg2017flownet}. By further improving the architectures and training strategies, deep learning based methods \cite{ilg2017flownet,sun2018pwc,sun2019models,yang2019volumetric,teed2020raft} have demonstrated stronger performance and faster inference speed compared with traditional optimization based approaches \cite{horn1981determining,zach2007duality,sun2010secrets,revaud2015epicflow,yang2015dense,brox2010large}.

\begin{figure}
	\centering
	\includegraphics[width=0.98\linewidth]{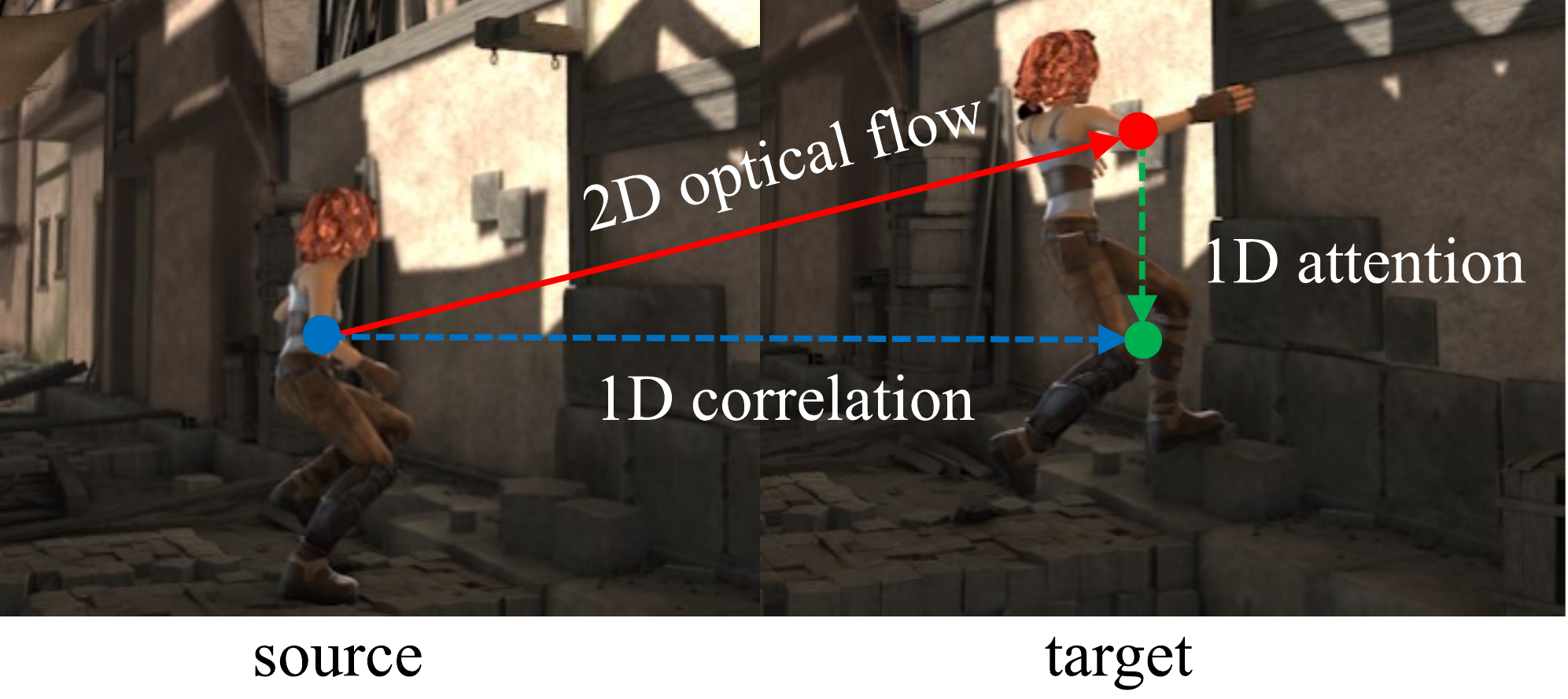}
	\vspace{-8pt}
	\caption{Optical flow factorization. We factorize the 2D optical flow with 1D attention and correlation in orthogonal directions to achieve large displacements search on high-resolution images. Specifically, for the correspondence (\textcolor{red}{red} point) of the \textcolor{blue}{blue} point, we first perform a 1D vertical attention to propagate the information of the red point to the \textcolor{green}{green} point, which lies on the same row of the blue point. Then a simple 1D correlation in the horizontal direction can be applied to build a horizontal cost volume. The vertical cost volume can be derived likewise with switched attention and correlation directions.}
	\label{fig:cost_volume_illustration}
	\vspace{-8pt}
\end{figure}

An essential component in deep learning based optical flow frameworks is \emph{cost volume} \cite{hosni2012fast}, which is usually computed by the dot product operation (also known as \emph{correlation} \cite{dosovitskiy2015flownet}) between two feature vectors. It stores the matching costs between each pixel in the source image and its potential correspondence candidates in the target image. By explicitly constructing a cost volume layer that encodes the search space, the network learns to better reason about the relative pixel displacements, as demonstrated by the superior performance of FlowNetC than FlowNetS that without such a layer \cite{dosovitskiy2015flownet,ilg2017flownet}.

The original cost volume in FlowNetC \cite{dosovitskiy2015flownet} is constructed in a single scale and it has difficulty in modeling large displacements due to the quadratic complexities with respect to the search window. PWC-Net \cite{sun2018pwc} migrates this problem by constructing multiple partial cost volumes in a coarse-to-fine warping framework. However, coarse-to-fine methods tend to miss small objects since they might not be visible in the highly-downsampled coarse scales, and thus have little chance to be correctly estimated \cite{revaud2015epicflow,steinbrucker2009large,brox2010large}. Moreover, warping might introduce artifacts in occlusion regions \cite{lu2020devon}, which may potentially hinder the network to learn correct correspondences. 

Current state-of-the-art optical flow method, RAFT \cite{teed2020raft}, maintains a single-resolution feature map and gradually estimates the flow updates in an iterative manner, eliminating several limitations of previous coarse-to-fine frameworks. One key component in RAFT is a 4D cost volume ($H\!\times\!W\!\times\!H\!\times\!W$) that is obtained by computing the correlations of all pairs. Thanks to such a large cost volume, RAFT achieves striking performance on established benchmarks. Nevertheless, the 4D cost volume requirement makes it difficult to scale to very high-resolution inputs due to the quadratic complexity with respect to the image resolution. Although one can partially alleviate this problem by processing on downsampled images, some fine-grained details, which might be critical for some scenarios 
(\eg, ball sports and self-driving cars), 
will be inevitably lost in such a process.
Furthermore, with the popularity of consumer-level high-definition cameras, it is much easier than before to get access to high-resolution videos, which accordingly raises the demand to be able to process such high-resolution videos with high efficiency.

To this end, we propose a new cost volume construction method for high-resolution optical flow estimation. Our key idea is to factorize the 2D search to two 1D substitutes in the vertical and horizontal direction, respectively, such that we can use 1D correlations to construct compact cost volumes.

Intuitively, such 1D correlations are not sufficient for optical flow estimation which is inherently a 2D search problem. However, as illustrated in Fig.~\ref{fig:cost_volume_illustration}, if we can propagate the information on the target image along the direction orthogonal to the correlation direction, the computed cost volume will contain meaningful correlation between the source pixel and its correspondence. This insight motivates us to design proper feature propagation and aggregation schemes for 1D correlation. Inspired by Transformers~\cite{vaswani2017attention}, we propose to learn such propagation with the attention mechanism, where we first apply 1D self attention on source feature (not shown in Fig.~\ref{fig:cost_volume_illustration} for brevity), then 1D cross attention between the source and target features (see Fig.~\ref{fig:self_cross_attn}).

Our 1D formulation yields two 3D cost volumes of size ($H\!\times\!W\!\times\!W$) and ($H\!\times\!W\!\times\!H$), respectively, which are then concatenated for the subsequent optical flow regression. This way, we reduce the complexity of all-pair correlation \cite{teed2020raft} from $\mathcal{O}(H\!\times\!W\!\times\!H\!\times W)$ to $\mathcal{O} (H\!\times\!W\!\times\!(H\!+\!W))$, enabling our method to scale to very high-resolution inputs with significant less computation. For example, our method consumes $6\times$ less memory than RAFT on 1080p ($1080 \times 1920$) videos. We also show flow results on real-world 4K ($2160 \times 3840$) resolution images and our method can handle images more than 8K ($4320 \times 7680$) resolution on a GPU with 32GB memory. Meanwhile, the evaluation on Sintel \cite{butler2012naturalistic} and KITTI \cite{menze2015object} shows that the accuracy of our method is only slightly worse than RAFT but outperforms other methods such as FlowNet2~\cite{ilg2017flownet} and PWC-Net~\cite{sun2018pwc}.

\textbf{Our contributions} can be summarized as follows:
\begin{itemize}
	\vspace{-5pt}
	\item We explore an innovative cost volume construction method which is fundamentally different from all existing methods. 
	\vspace{-5pt}
	\item We show that cost volumes constructed using 1D correlations, despite somewhat counter-intuitive, can achieve striking flow estimation accuracy comparable to the state of the art.
	\vspace{-5pt}
	\item Our method is slightly inferior compared to RAFT in terms of accuracy but enjoys significantly less memory consumption, enabling us to process very high-resolution images (more than 8K resolution ($4320 \times 7680$) in our experiment).
\end{itemize}

\section{Related Work}

Optical flow has traditionally been formulated as an optimization problem \cite{horn1981determining,brox2010large}. A thorough comparison and evaluation of traditional methods can be found at \cite{sun2010secrets}. In this section, we mainly review recent learning based methods from different perspectives.

\paragraph{Cost volume.} The concept of cost volume dates back to stereo matching literature \cite{hosni2012fast,scharstein2002taxonomy}. A cost volume stores the matching costs for different pixel displacement candidates (\ie, disparity in stereo) at each pixel coordinate. Thus it is a 3D tensor ($H \times W \times D$) in stereo matching, where $D$ is the maximum disparity range. The cost volume serves as a discriminative representation of the search space and powerful optimization methods can be employed to filter outliers, and thus it usually leads to accurate results \cite{hirschmuller2007stereo,hosni2012fast}. 

The effectiveness of cost volume also benefits the optical flow community, either traditional \cite{chen2016full,xu2017accurate} or learning based methods \cite{dosovitskiy2015flownet,sun2018pwc,yang2019volumetric,lu2020devon,zhao2020maskflownet,teed2020raft}. However, unlike 1D disparity in stereo, the search space in optical flow is 2D, thus resulting in a 4D cost volume ($H \times W \times (2R+1) \times (2R + 1)$) for search radius $R$, which is computational expensive for large displacements. To alleviate this issue, a popular strategy is to use coarse-to-fine warping schemes \cite{brox2004high,sun2018pwc,yang2019volumetric,zhao2020maskflownet}. However, coarse-to-fine methods tend to miss fast-moving small objects \cite{revaud2015epicflow,lu2020devon,teed2020raft}. Recently, RAFT \cite{teed2020raft} proposes to construct a 4D cost volume ($H \times W \times H \times W$) by computing all spatial correlations. Despite the state-of-the-art performance, RAFT is inherently constrained by the input resolution due to the quadratic complexity. 
In contrast, we propose to factorize the 2D optical flow with 1D attention and correlation, leading to significantly reduced complexities and allowing us to handle larger image resolutions.

\begin{figure*}
	\centering
	\includegraphics[width=\linewidth]{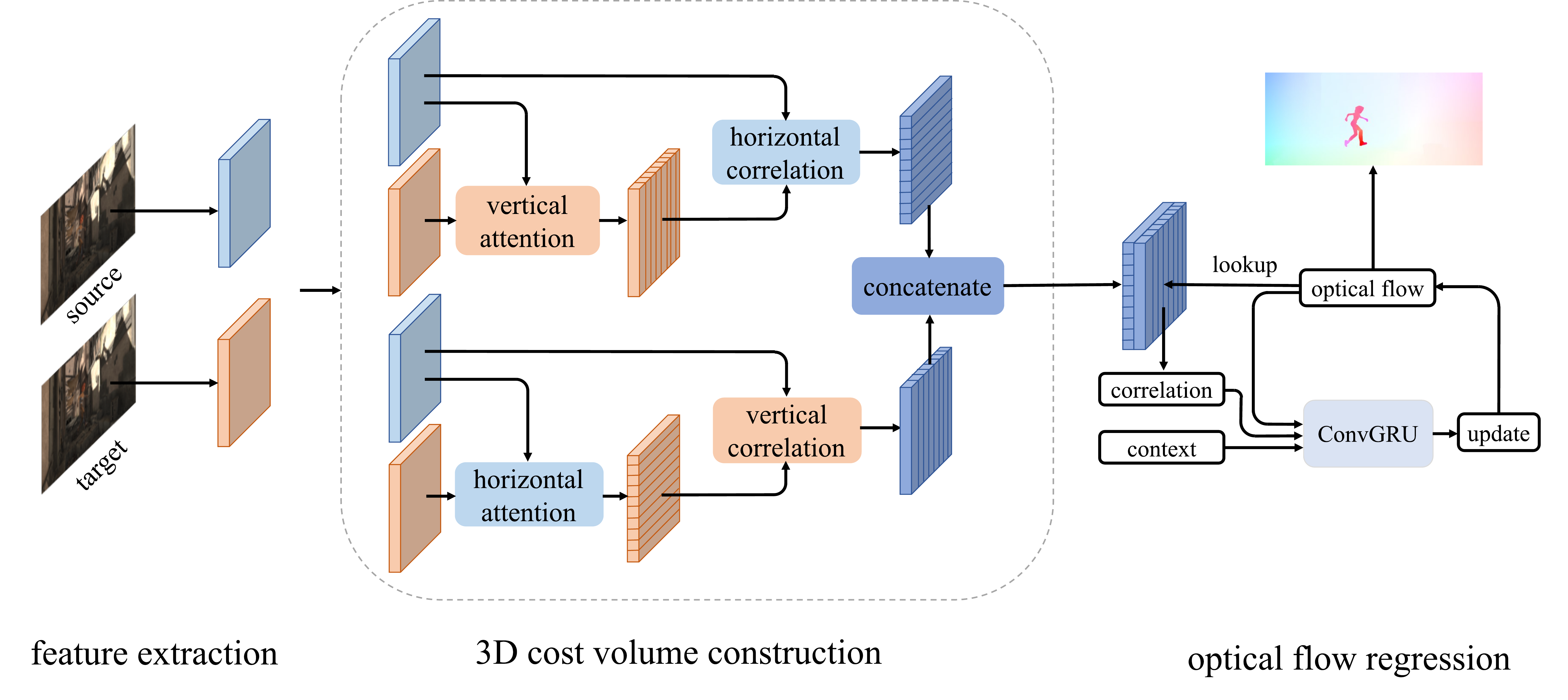}
	\vspace{-22pt}
	\caption{Overview of our framework. Given a pair of source and target images, we first extract $8 \times $ downsampled features with a shared backbone network. The features are then used to construct two 3D cost volumes with \emph{vertical attention, horizontal correlation} and \emph{horizontal attention, vertical correlation}, respectively. The two cost volumes are then concatenated for optical flow regression, where we adopt RAFT's framework to estimate the optical flow in an iterative manner. After a number of iterations, the final optical flow prediction can be obtained. More details can be found in Sec.~\ref{sec:framework}.} 
	\label{fig:overview}
	\vspace{-6pt}
\end{figure*}

\paragraph{Flow regression scheme.} Top-performing learning based optical flow methods mostly share a similar structure: starting from a coarse flow estimate and then gradually refining the initial prediction \cite{hur2019iterative}. They can be broadly classified into two categories: \emph{non-iterative} and \emph{iterative} methods. Here, we use `non-iterative' to denote that the refinement networks have their own trainable weights, while `iterative' represents the network weights are shared at each refinement stage. Representative non-iterative frameworks include FlowNet2~\cite{ilg2017flownet}, PWC-Net~\cite{sun2018pwc} and their variants \cite{hui2018liteflownet,yin2019hierarchical,yang2019volumetric,zhao2020maskflownet}. IRR-PWC \cite{hur2019iterative} and RAFT \cite{teed2020raft} are two representative iterative methods. IRR-PWC shares a convolutional decoder to estimate flow in a coarse-to-fine framework with a limited number of iterations, while RAFT uses a ConvGRU \cite{cho2014properties} and applies this architecture for a large number of iterations ($10+$). In this paper, we validate the effectiveness of our proposed cost volume construction method with RAFT's framework due to its compactness and good performance. But in theory, our key idea is orthogonal to the flow regression scheme adopted.

\paragraph{Attention mechanism.} 
Attention mechanism has achieved remarkable success in modeling long-range dependencies. However, it has one crucial limitation that the computational complexity grows quadratically with respect to the input size. This issue becomes severer when applying attention to vision tasks due to the large number of image pixels. Large volumes of work tries to reduce the complexity of attention by sparse connection patterns \cite{child2019generating}, low-rank approximations \cite{wang2020linformer} or recurrent operations \cite{dai2019transformer}. A thorough review of efficient attention mechanisms can be found at \cite{tay2020efficient}. Among these methods, perhaps the most relevant to ours in vision are CCNet~\cite{huang2019ccnet} and Axial-Deeplab~\cite{wang2020axial}, both of which use two 1D self attentions for global dependency modeling. Different from the self attention in CCNet and Axial-Deeplab, we use 1D cross attention and 1D correlation between a pair of source and target images to achieve large displacement correspondence search. Moreover, the output is a cost volume in our case to explicitly model matching costs while CCNet and Axial-Deeplab output a feature map.

\section{Optical Flow Factorization}
\label{sec:cost_volume}

Optical flow is inherently a 2D search problem, but directly searching on the 2D image space is computationally intractable for very large displacements due to the quadratic complexity with respect to the search window. For example, the potential search space can be up to $10^4$ pixels for search range $[-50, 50]$. 
This problem is yet more pronounced for high-resolution images.

To enable optical flow estimation on high-resolution images, we observe that the 2D search space can be approximated in two 1D directions. As illustrated in Fig.~\ref{fig:cost_volume_illustration}, to achieve 2D correspondence modeling effect between the blue and red points, we can first propagate the information of the red point to the green one, which lies on the same row of the blue point. Then a 1D search along the horizontal direction can capture the information of the red point.

In light of this, we propose to factorize the 2D optical flow with 1D attention \cite{vaswani2017attention} and 1D correlation \cite{dosovitskiy2015flownet} in orthogonal directions to achieve large displacements search on high-resolution images. Specifically, we first perform a 1D attention in the vertical direction, which computes a weighted combination of pixels that are in the same column. Then a simple 1D correlation in the horizontal direction can achieve 2D search effect. The resulting cost volume is 3D, similar to stereo methods \cite{mayer2016large,xu2020aanet}. By exchanging the direction of attention and correlation, we can obtain another 3D cost volume. These two cost volumes are concatenated for optical flow regression (see Fig.~\ref{fig:overview}).

Next, we present details of our \emph{horizontal} cost volume construction with 1D vertical attention and 1D horizontal correlation. 
Its vertical counterpart can be similarly derived.
For source and target images ${\bm I}_1$ and ${\bm I}_2$, a shared convolutional backbone is first used to extract features (see Fig.~\ref{fig:overview}), we then construct the cost volume at the feature level.

\begin{figure}
	\centering
	\includegraphics[width=0.98\linewidth]{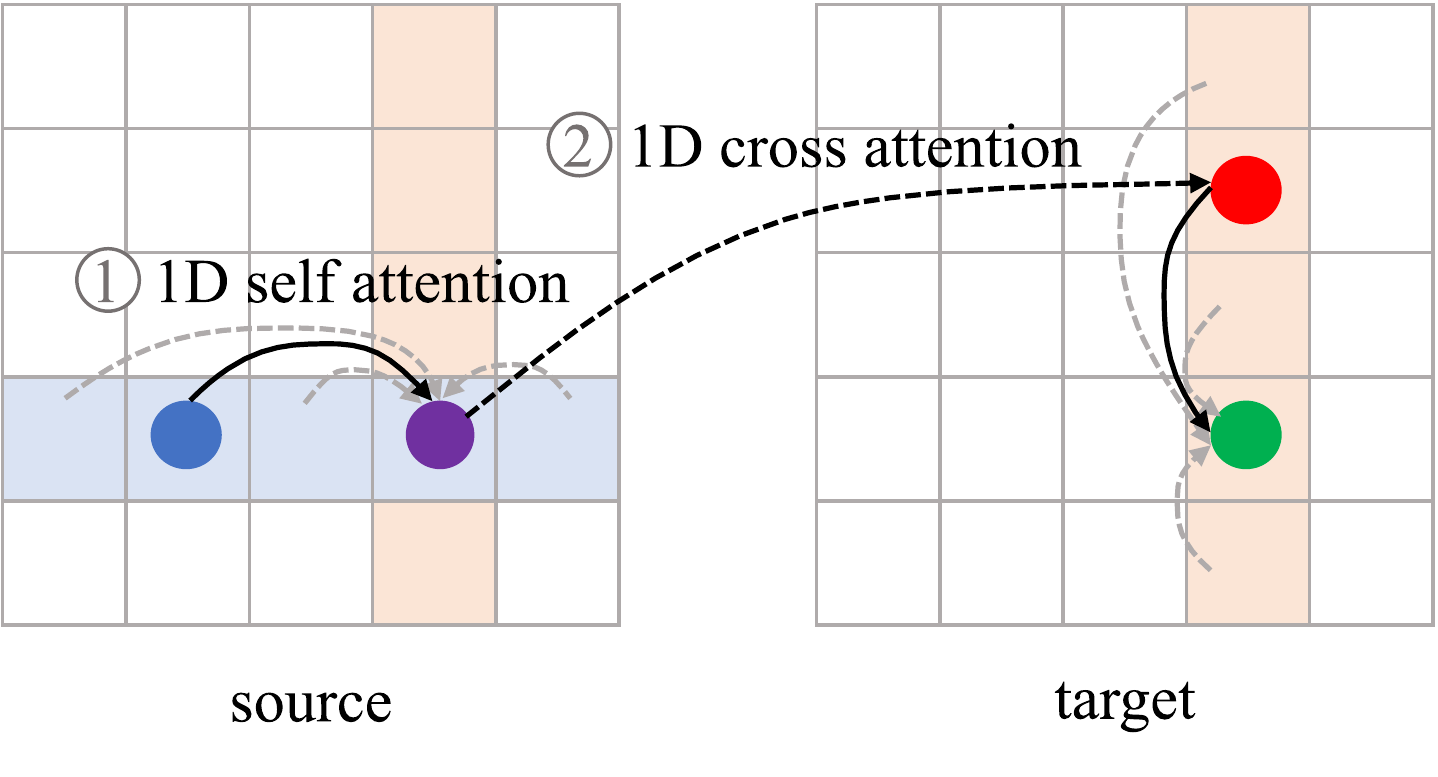}
	\vspace{-12pt}
	\caption{Illustration of self and cross attentions in the vertical aggregation process of target image feature. To properly propagate the information of \textcolor{red}{red} point (\textcolor[RGB]{68,114,196}{blue} point's correspondence) onto the \textcolor[RGB]{0,176,80}{green} point, we first perform a 1D horizontal self attention on the source image to make every point encode the information of its entire row (thus the \textcolor[RGB]{112,48,160}{purple} point is aware of the {blue} point). Then a 1D vertical cross attention between the source and target is performed, where the {green} point receives the information of the {red} point \emph{conditioned} on the {purple} point feature. %
	}
	\label{fig:self_cross_attn}
	\vspace{-6pt}
\end{figure}

\begin{figure}
	\centering
	\includegraphics[width=0.98\linewidth]{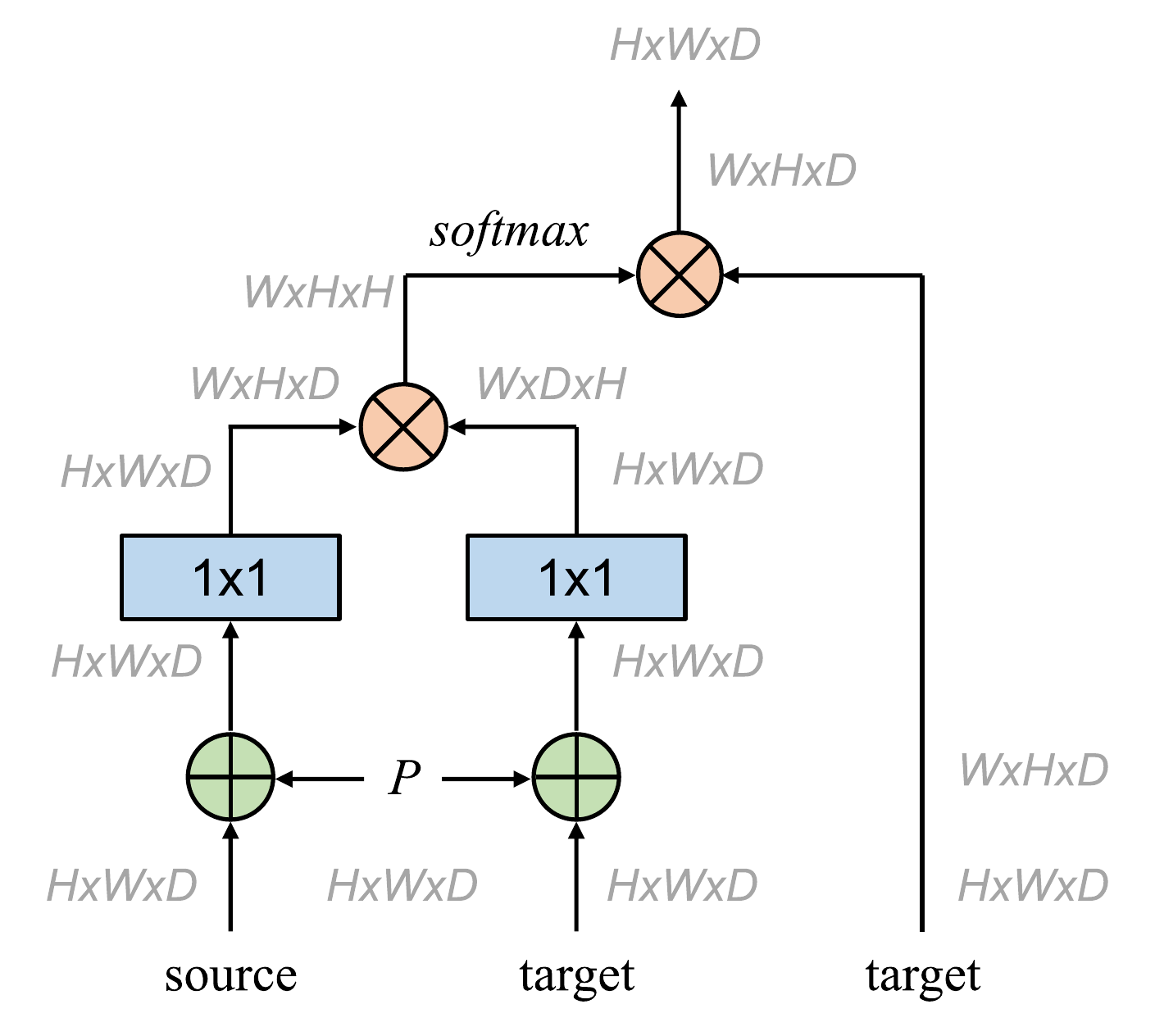}
	\vspace{-12pt}
	\caption{Computational graph of 1D cross attention in vertical direction. The positional encoding ${\bm P}$ is added to the input features. Two $1 \times 1$ convolutions are used to learn the attention weights. ``$\bigoplus$'' denotes element-wise addition and ``$\bigotimes$'' denotes matrix multiplication on the last two dimensions. The softmax operation is performed on the last dimension. The matrix dimensions are indicated with grey font, proper reshaping is performed when noted.
	}
	\label{fig:cross_attn}
	\vspace{-6pt}
\end{figure}

\subsection{1D Attention}
\label{sec:1d_attn}

With source and target features ${\bm F}_1, {\bm F}_2 \in \mathbb{R}^{H \times W \times D}$, where $H, W, D$ represents height, width and feature dimension, respectively, our goal is to generate a new feature $\hat{{\bm F}}_2$ where every feature vector is aware of the features of points in ${\bm F}_2$ lying in the same column. In this way, we can then perform a simple 1D search along the horizontal direction to achieve 2D correspondence modeling effect.

Concretely, we define $\hat{{\bm F}}_2$ as a linear combination of column-wise features from ${\bm F}_2$:
\begin{equation}
\begin{split}
\label{eq:linear}
\hat{{\bm F}}_2 (h, w) & = \sum_{i = 0}^{H-1} f_{i, h, w}  {\bm F}_2 (i, w), \\
\end{split}
\end{equation}
where $h = 0, 1, \cdots, H-1, w = 0, 1, \cdots, W-1$, $\hat{{\bm F}}_2 (h, w)$ is the output feature vector at position $(h, w)$, $f_{i, h, w}$ is the combination weight. The key in our formulation lies in how to find $f_{i, h, w}$ that can properly aggregate the column-wise information to aid in the subsequent 1D correspondence search (Fig.~\ref{fig:cost_volume_illustration}). 
Inspired by the success of Transformers~\cite{vaswani2017attention} in modeling long-range dependencies, we propose to learn the combination coefficients with the attention mechanism.

However, different from the original attention that computes all pair-wise similarities, we only perform 1D attention in an axis-aligned manner.
As illustrated in Fig.~\ref{fig:self_cross_attn}, the goal of the attention operation is to propagate the information of the red point (blue point's correspondence) onto the green one, which lies on the same row of the blue point. To achieve this, one na\"ive solution is applying a 1D cross attention operation between the same column of the source and target, to make the target feature aggregation dependent on the source feature. However, the same column in the source image may not contain the corresponding pixel (blue point), rendering it difficult to learn proper aggregation. To resolve this issue, we first perform a 1D self attention operation in the horizontal direction of the source feature before computing the cross attention, which propagates the information of the corresponding point on source image to the entire row.

The computational graph of 1D vertical cross attention is illustrated in Fig.~\ref{fig:cross_attn}. The inputs are self-attended source feature and original target feature. We also introduce position information ${\bm P} \in \mathbb{R}^{H \times W \times D}$ in the attention computation where ${\bm P}$ is a fixed 2D positional encoding same as DETR \cite{carion2020end}. We first use two $1 \times 1$ convolutions to project ${\bm F}_1 + {\bm P}$ and ${\bm F}_2 + {\bm P}$ into the embedding space $\Tilde{{\bm F}}_1$ and $\Tilde{{\bm F}}_2$. Then the attention matrix on the vertical direction can be obtained by first reshaping $\Tilde{{\bm F}}_1$ and $\Tilde{{\bm F}}_2$ to $W\times H \times D$ and $W \times D \times H$, respectively, and then performing a matrix multiplication on the last two dimensions, which results in a $W \times H \times H$ matrix. By normalizing the last dimension with the softmax function, we obtain the attention matrix ${\bm A} \in \mathbb{R}^{W \times H \times H}$. The final attended feature $\hat{{\bm F}}_2$ can be computed by multiplying matrix ${\bm A}$ with the reshaped target feature ($W \times H \times D$) on the last two dimensions. By reshaping the resulting $W \times H \times D$ matrix to $H \times W \times D$, we obtain the final feature $\hat{{\bm F}}_2$. After the 1D cross attention operation, each position in $\hat{{\bm F}}_2$ has encoded the information of positions that are in the same column. This process can be similarly applied to 1D horizontal self attention computation by replacing the target features with source features and performing matrix multiplication on the width dimension.

\subsection{1D Correlation}

With vertically aggregated feature $\hat{{\bm F}}_2$, we can perform a simple 1D search along the horizontal direction to construct a 3D cost volume ${\bm C} \in \mathbb{R}^{H \times W \times (2R + 1)}$, similar to stereo methods \cite{mayer2016large,xu2020aanet}. We use $R$ to represent the search radius along the horizontal direction, and we have
\begin{equation}
\label{eq:1d_corr}
{\bm C} (h, w, R\!+\!r) =\frac{1}{\sqrt{D}} {\bm F}_1(h, w) \boldsymbol{\cdot} \hat{{\bm F}}_2(h, w\!+\!r)
\end{equation}
where $\boldsymbol{\cdot}$ denotes the dot-product operator, $r\!\in\!\{ -R, -R + 1, \cdots, 0, \cdots, R - 1, R\}$, and $\frac{1}{\sqrt{D}}$ is a normalization factor to avoid large values after dot product following \cite{vaswani2017attention}. In our implementation, we pre-compute a 3D cost volume of size $H \times W \times W$ by performing matrix multiplication between ${\bm F}_1$ and $\hat{{\bm F}}_2$ on the width dimension. Then the equivalent form of Eq.~\eqref{eq:1d_corr} can be obtained by performing a lookup operation on the 3D cost volume with search radius $R$.

Although we only perform 1D correlations, our method can model 2D correspondence thanks to the attention operation. Specifically, the theoretical search range of our cost volume construction method is $(2R+1)(H + W) - (2R+1)^2$. This can be proven by substituting Eq.~\eqref{eq:linear} into Eq.~\eqref{eq:1d_corr}: 
\begin{equation}
\begin{split}
\label{eq:correlation}
{\bm C} (h, w, R\!+\!r) 
& \!=\! \frac{1}{\sqrt{D}} {\bm F}_1(h, w) \!\boldsymbol{\cdot}\! \sum_{i = 0}^{H-1}\!f_{i, h, w + r}  {\bm F}_2 (i, w\!+\!r) \\
& \!=\! \frac{1}{\sqrt{D}}\!\sum_{i = 0}^{H-1}\!f_{i, h, w+r}  [{\bm F}_1(h, w)\!\boldsymbol{\cdot}\!{\bm F}_2 (i, w\!+\!r)], \\
\end{split}
\end{equation}
where $f_{i, h, w+r}$ is defined with the 1D attention operation in Sec.~\ref{sec:1d_attn}. From term ${\bm F}_1(h, w) \boldsymbol{\cdot} {\bm F}_2 (i, w + r)$ in Eq.~\eqref{eq:correlation} with $i = 0, 1, \cdots H-1$, and $r = -R, -R + 1, \cdots, 0, \cdots, R - 1, R$, we can see that the search range of position $(h, w)$ spans vertically to the image height and horizontally to the maximum search radius $R$. Therefore, the theoretical search range for vertical attention and horizontal correlation is $H(2R+1)$.
Similarly, when performing horizontal attention and vertical correlation, we can obtain another cost volume $\Tilde{{\bm C}} \in \mathbb{R}^{ H \times W \times (2R + 1)}$, where
\begin{equation}
\Tilde{{\bm C}} (h, w, R + r)\!=\!\frac{1}{\sqrt{D}} \sum_{j = 0}^{W-1}\!f_{j, h + r, w}  [{\bm F}_1(h, w) \boldsymbol{\cdot} {\bm F}_2 (h + r, j)].
\end{equation}
In this case, the theoretical search range is $W(2R+1)$. 
Concatenating these two cost volumes results in a cost volume of shape $H \times W \times 2(2R+1)$, and the theoretical search range becomes $(2R+1)(H + W) - (2R+1)^2$, where $(2R+1)^2$ is the overlapping area of these two cost volumes. 

As a comparison, previous local window based approaches \cite{dosovitskiy2015flownet,ilg2017flownet,sun2018pwc} usually construct a cost volume of shape $H \times W \times (2R+1)^2$ and the search range is $(2R+1)^2$. Thus, our method enjoys a larger theoretical search range while maintaining a much smaller cost volume.

\begin{figure*}[t]
    \centering
    \begin{subfigure}{0.19\linewidth}
        \includegraphics[width=\linewidth]{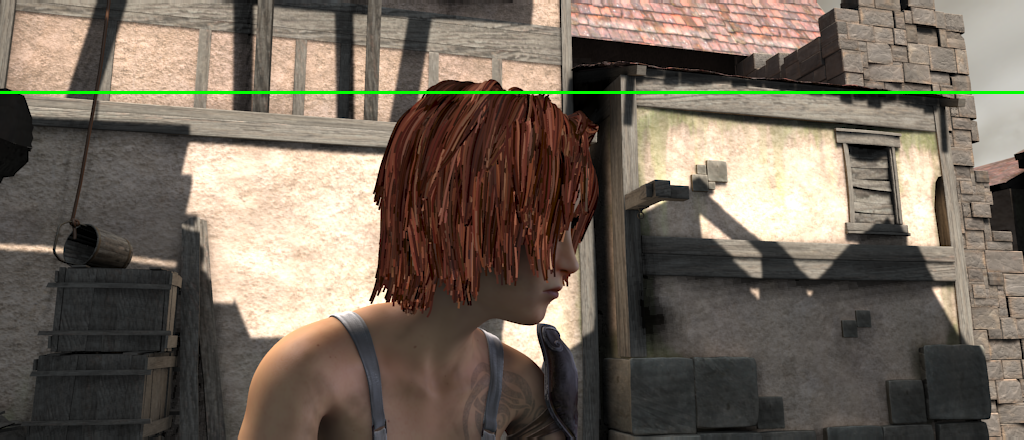}
        \vspace{-16pt}
        \caption{\small{source image}}
        \label{fig:source}
    \end{subfigure}
    \begin{subfigure}{0.19\linewidth}
        \includegraphics[width=\linewidth]{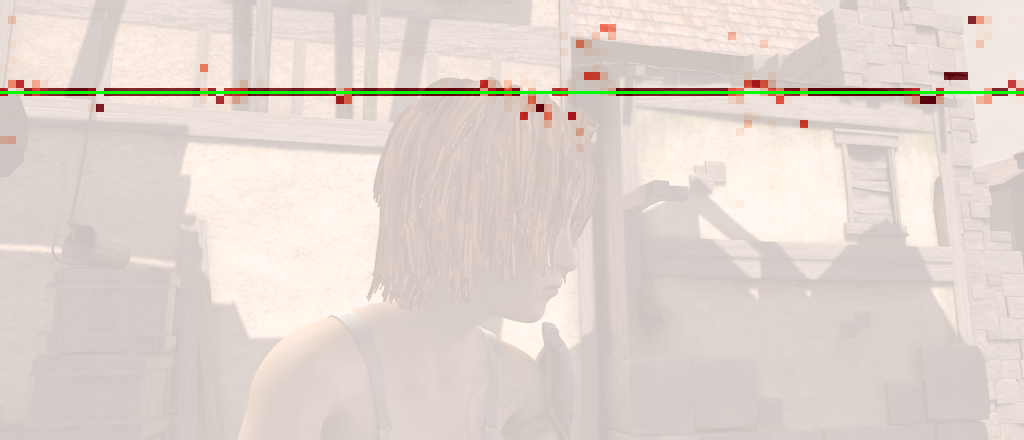}
        \vspace{-16pt}
        \caption{\small{frame stride 1}}
        \label{fig:stride1}
    \end{subfigure}
    \begin{subfigure}{0.19\linewidth}
        \includegraphics[width=\linewidth]{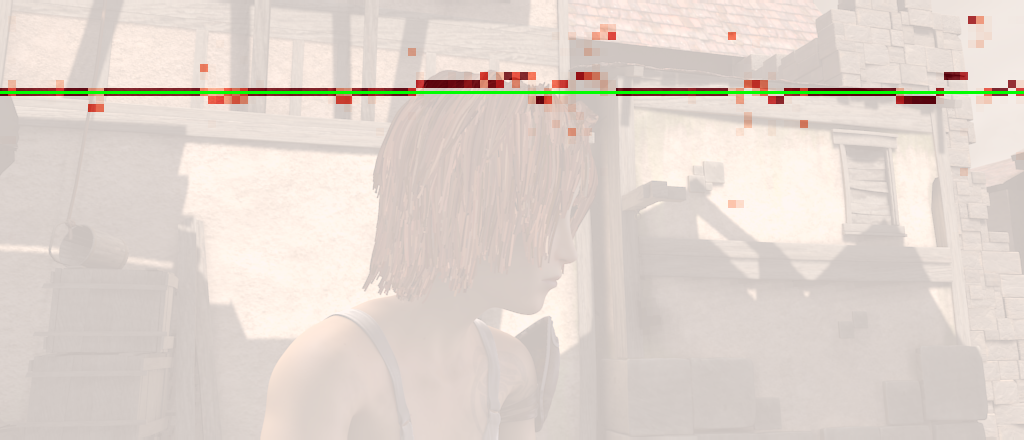}
        \vspace{-16pt}
        \caption{\small{frame stride 3}}
        \label{fig:stride3}
    \end{subfigure}
    \begin{subfigure}{0.19\linewidth}
        \includegraphics[width=\linewidth]{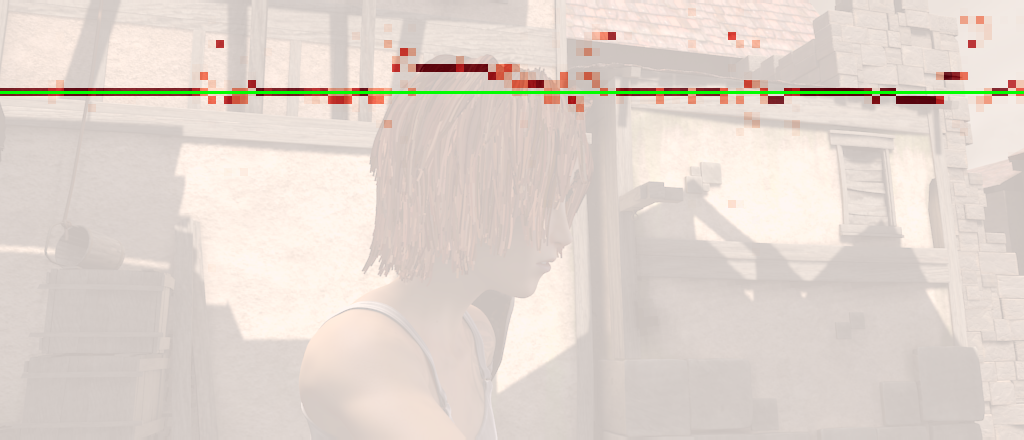}
        \vspace{-16pt}
        \caption{\small{frame stride 6}}
        \label{fig:stride6}
    \end{subfigure}
    \begin{subfigure}{0.19\linewidth}
        \includegraphics[width=\linewidth]{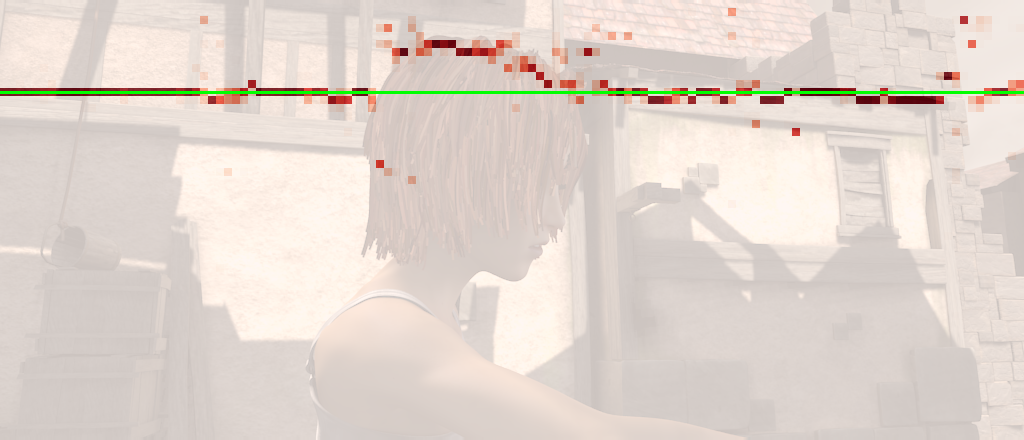}
        \vspace{-16pt}
        \caption{\small{frame stride 9}}
        \label{fig:stride9}
    \end{subfigure}
    
    \vspace{-8pt}
    \caption{Visualization of 1D \emph{vertical} attentions.
    We fix the source image (\ref{fig:source}) and select target frames (\ref{fig:stride1}-\ref{fig:stride9}) from a video sequence with gradually increased frame strides. The attention weights for the \textcolor{green}{green} row pixels are visualized onto the image, which shows the attentions generally focus on relevant pixels (\eg, the moving head). 
    }
    \label{fig:vis_attn_y}
    \vspace{-4pt}
\end{figure*}

\begin{figure*}[t]
    \centering
    \begin{subfigure}{0.24\linewidth}
        \includegraphics[width=\linewidth]{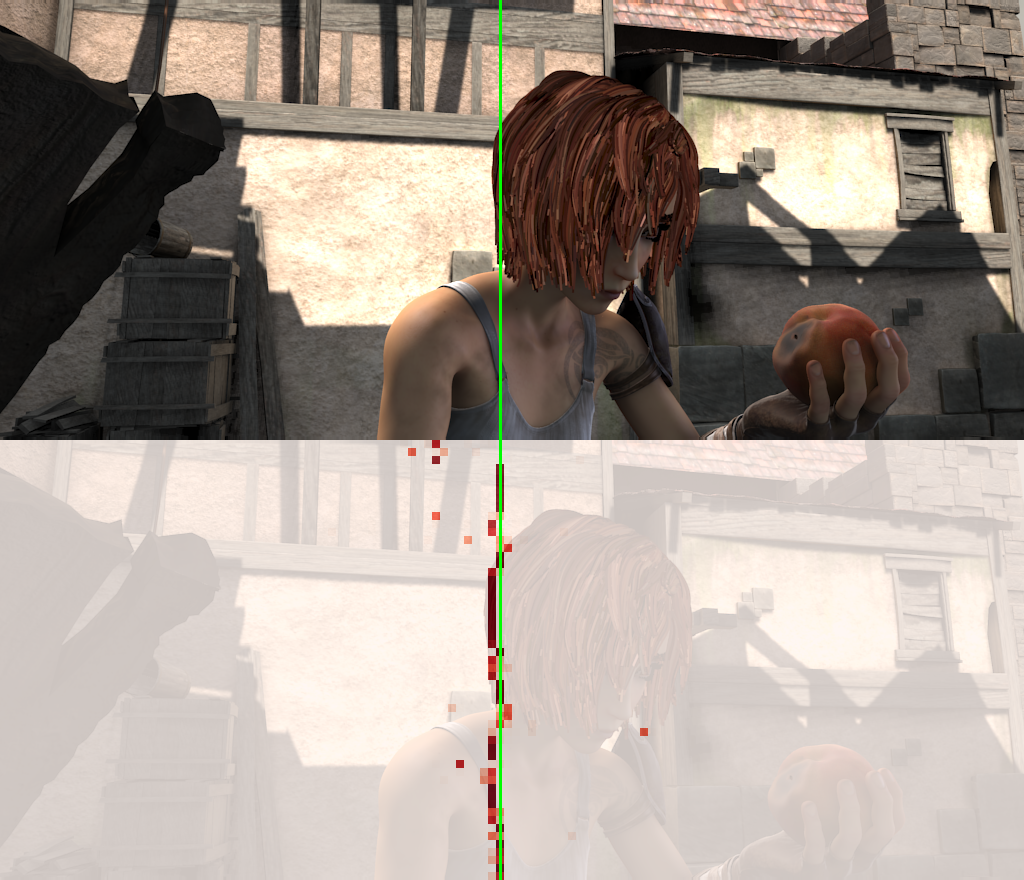}
        \vspace{-16pt}
        \caption{\small{frame stride 1}}
        \label{fig:stride1_x}
    \end{subfigure}
    \begin{subfigure}{0.24\linewidth}
        \includegraphics[width=\linewidth]{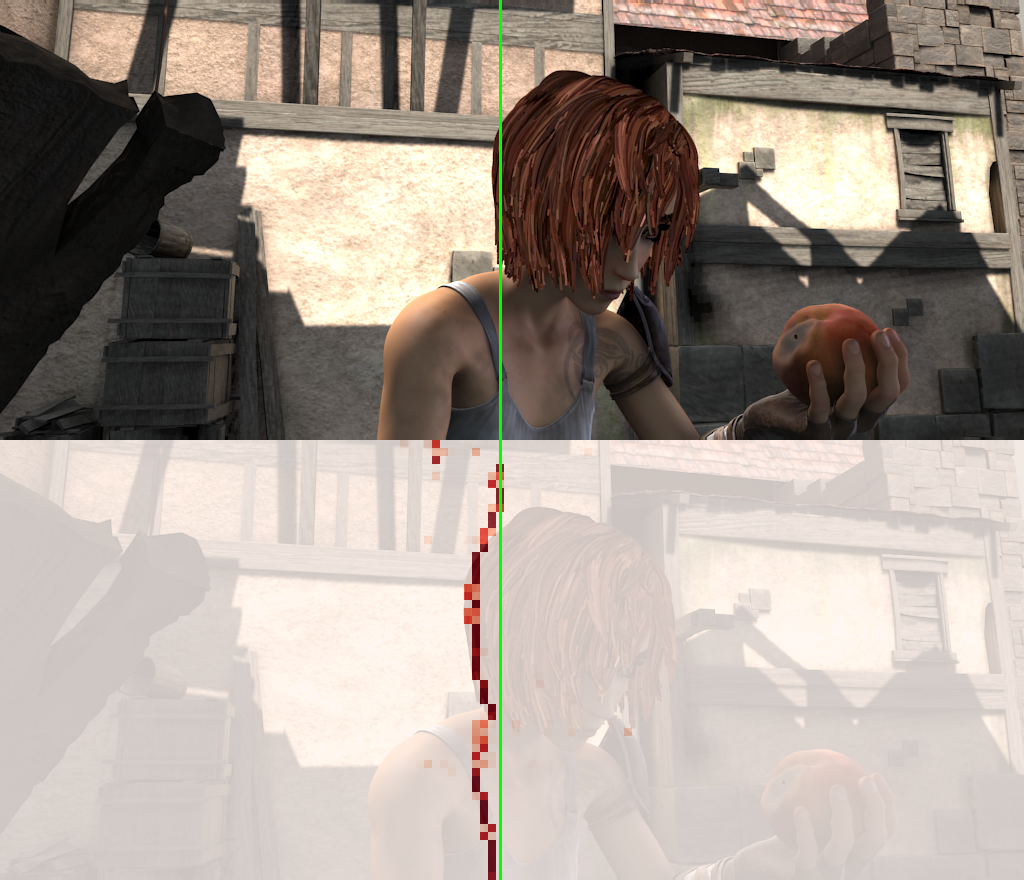}
        \vspace{-16pt}
        \caption{\small{frame stride 3}}
        \label{fig:stride3_x}
    \end{subfigure}
    \begin{subfigure}{0.24\linewidth}
        \includegraphics[width=\linewidth]{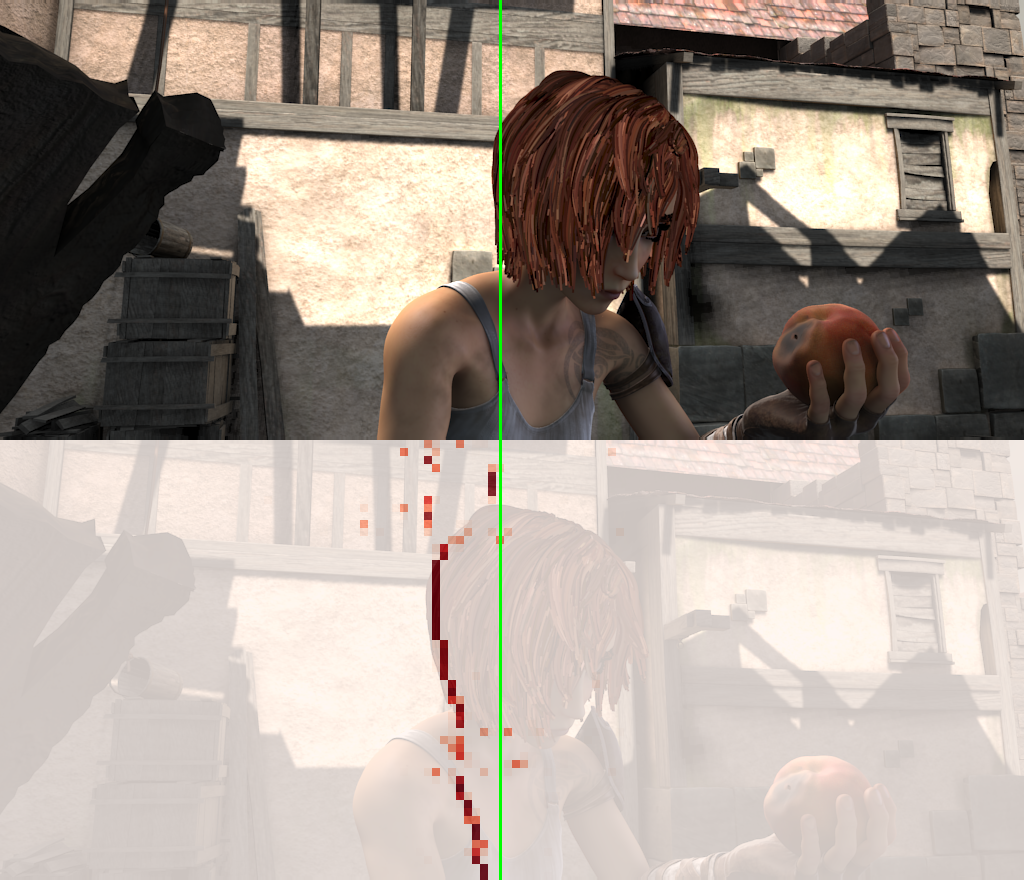}
        \vspace{-16pt}
        \caption{\small{frame stride 6}}
        \label{fig:stride6_x}
    \end{subfigure}
    \begin{subfigure}{0.24\linewidth}
        \includegraphics[width=\linewidth]{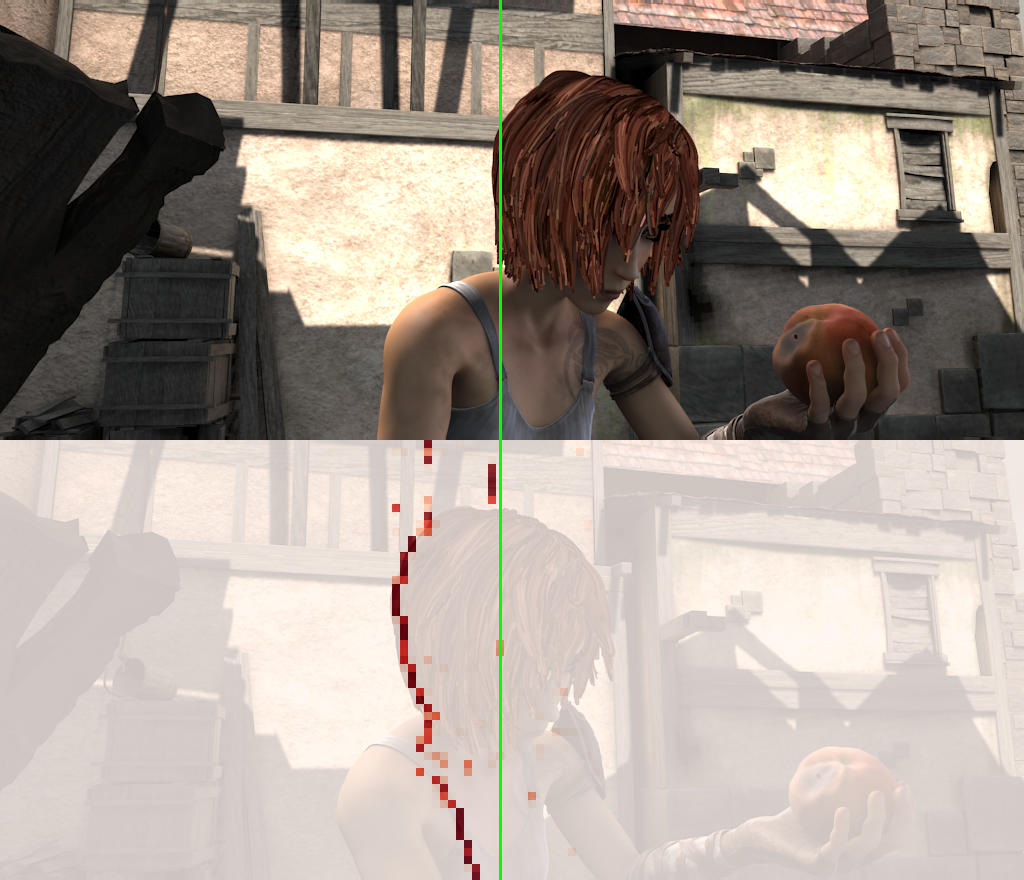}
        \vspace{-16pt}
        \caption{\small{frame stride 9}}
        \label{fig:stride9_x}
    \end{subfigure}

    \vspace{-8pt}
    \caption{Visualization of 1D \emph{horizontal} attentions. Similar phenomenon can be observed as Fig.~\ref{fig:vis_attn_y}.
    }
    \label{fig:vis_attn_x}
    \vspace{-8pt}
\end{figure*}

\section{Flow Regression Framework}
\label{sec:framework}

We verify the effectiveness of our proposed cost volume construction method with RAFT framework \cite{teed2020raft}. As illustrated in Fig.~\ref{fig:overview}, we first extract $8 \times$ downsampled features for source and target images, and then construct two 3D cost volumes with our proposed method in Sec.~\ref{sec:cost_volume}. Next, a shared update operator is applied iteratively to estimate the flow updates. At each iteration, a set of correlation values will be generated by looking up the 3D cost volume with current flow estimate. A context feature is also extracted from the source image with an additional network following RAFT, which is omitted in Fig.~\ref{fig:overview} for brevity. The correlations, together with the estimated flow and context feature, are then fed to a ConvGRU unit to produce a flow update, which is added to the current flow estimate.

In our framework, the lookup operation is defined by indexing the pre-computed cost volume with the current flow estimate. Specifically, if the current flow estimate is ${\bm f} = (f_x, f_y)$, then the lookup center becomes $(h + f_y, w + f_x)$ for pixel position $(h, w)$. We index the 3D cost volumes both horizontally and vertically within a search range of $R$ and obtain two 3D cost volumes ($H \times W \times (2R + 1)$) that are concatenated for optical flow regression. We note that the correlation lookup is equivalent to Eq.~\eqref{eq:1d_corr} by modifying the correlation center to $(h + f_y, w + f_x)$.

\begin{table}[t]
    \centering

    \begin{tabular}{lccc}
    \toprule
    \multirow{2}{*}[-2pt]{Cost volume} & \multicolumn{3}{c}{Sintel (train, clean)} \\
    \addlinespace[-12pt] \\
    \cmidrule(lr){2-4}
    \addlinespace[-12pt] \\
    & EPE & EPE ($x$) & EPE ($y$) \\
    \midrule
    $y$ attn, $x$ corr & 3.10 & 1.66 & 2.12 \\
    $x$ attn, $y$ corr & 4.05 & 3.55 & 1.13 \\

    concat both & {\bf 1.98} & {\bf 1.48} & {\bf 0.94} \\
    
    \bottomrule
    \end{tabular}
    \caption{Analysis on horizontal ($x$) and vertical ($y$) cost volumes. EPE ($x$) and EPE ($y$) represent the end-point-error of the horizontal and vertical flow component, respectively.}
    \label{tab:cost_volume_xy}
    \vspace{-6pt}
\end{table}

\section{Experiments}

\paragraph{Datasets and evaluation setup.} we consider two evaluation setups following previous methods \cite{ilg2017flownet,sun2018pwc,teed2020raft}. First, we pre-train our model on FlyingChairs \cite{dosovitskiy2015flownet} and FlyingThings3D \cite{mayer2016large} datasets, and then evaluate the cross-dataset generalization ability on Sintel \cite{butler2012naturalistic} and KITTI \cite{menze2015object} training sets. Second, we perform additional fine-tuning on Sintel and KITTI training sets and then evaluate on the online benchmarks. The end-point-error (EPE) is reported in evaluation. For KITTI, another evaluation metric, F1-all, which denotes percentage of outliers for all pixels, is also reported. For ablation study, we also use the EPE in different motion magnitudes to better understand the performance gains. Specifically, we use $s_{0-10}, s_{10-40}$ and $s_{40+}$ to denote the EPE over regions with speed in $0-10$, $10-40$ and more than $40$ pixels. For experiments on very high-resolution images, we mix FlyingThings3D, Sintel, HD1K \cite{kondermann2016hci} and Slow Flow \cite{janai2017slow} datasets for additional fine-tuning.

\paragraph{Implementation details.} We implement our framework in PyTorch \cite{paszke2019pytorch} and use AdamW \cite{loshchilov2017decoupled} as the optimizer. We follow RAFT \cite{teed2020raft} for dataset schedule and training hyper-paramters. Training is first conducted on FlyingChairs dataset for 100K iterations, followed by another 100K iterations on FlyingThings3D dataset. We also perform dataset-specific fine-tuning on Sintel and KITTI datasets. For training, we use 12 iterations for flow regression. For evaluation, the iteration numbers for Sintel and KITTI are set to 32 and 24, respectively. The search range $R$ in the cost volume lookup is set to $32$, which correspondences to $256$ pixels in the original image resolution. More implementation details are presented in \emph{supplementary materials}.

\begin{table*}[t]
    \centering

    \setlength{\tabcolsep}{3.pt} %
    \begin{tabular}{lccccccccccccc}
    \toprule
    \multirow{2}{*}[-2pt]{Method} & \multicolumn{4}{c}{Sintel (train, clean)} & \multicolumn{4}{c}{Sintel (train, final)} & \multicolumn{5}{c}{KITTI (train)} \\
    \addlinespace[-12pt] \\
    \cmidrule(lr){2-5} \cmidrule(lr){6-9} \cmidrule(lr){10-14}
    \addlinespace[-12pt] \\
    & EPE & $s_{0-10}$ & $s_{10-40}$ & $s_{40+}$ & EPE & $s_{0-10}$ & $s_{10-40}$ & $s_{40+}$ & EPE & $s_{0-10}$ & $s_{10-40}$ & $s_{40+}$  & F1-all \\
    \midrule
    w/o self \& cross & 2.94 & 1.75 & 4.65 & 24.96 & 4.72 & 2.43 & 6.69 & 30.27 & 14.01 & 0.99 & 3.87 & 29.06 & 38.25 \\
    w/o self & 2.15 & {\bf 1.44} & 3.83 & 20.87 & 3.54 & 2.16 & 5.49 & 25.29 & 8.67 & 0.89 & 2.49 & 17.82 & 27.25 \\
    w/o pos & \bf{1.98} & 1.49 & {\bf 3.80} & 19.65 & 3.61 & 2.49 & 5.61 & 25.45 & 7.71 & 0.84 & 2.31 & 15.49 & 25.37 \\
    Flow1D & \bf{1.98} & 1.68 & 3.92 & {\bf 18.86} & {\bf 3.27} & {\bf 2.06} & {\bf 5.18} & {\bf 23.76} & {\bf 6.69} & {\bf 0.80} & {\bf 2.18} & {\bf 13.41} & {\bf 22.95} \\
    \bottomrule
    \end{tabular}
    \vspace{-5pt}
    \caption{Ablation study for our cost volume design. Models are trained on FlyingChairs and FlyingThings3D.
    }
    \label{tab:ablation}
    \vspace{-10pt}
\end{table*}

\begin{table*}[t]
    \centering
    \begin{tabular}{lccccccccc}
    \toprule
    \multirow{2}{*}[-2pt]{Method} & \multicolumn{2}{c}{Sintel (train)} & \multicolumn{2}{c}{KITTI (train)} & \multirow{2}{*}[-2pt]{Params} &
    \multicolumn{2}{c}{$448\times 1024$} & \multicolumn{2}{c}{$1088\times 1920$} \\
    \addlinespace[-12pt] \\
    \cmidrule(lr){2-3} \cmidrule(lr){4-5} \cmidrule(lr){7-8} \cmidrule(lr){9-10}
    \addlinespace[-12pt] \\
    & Clean & Final & EPE  & F1-all & & Memory & Time (ms) & Memory & Time (ms) \\
    \midrule
    RAFT \cite{teed2020raft} & 1.43 & 2.71 & 5.04 & 17.40 & 5.26M & 0.48GB & 94 & 8.33GB & 393 \\
    FlowNet2 \cite{ilg2017flownet} & 2.02 & 3.14 & 10.06 & 30.37 & 162.52M & 1.31GB & 186 & 3.61GB & 496 \\
    PWC-Net \cite{sun2018pwc} & 2.55 & 3.93 & 10.35 & 33.67 & 9.37M & 0.86GB & 24 & 1.57GB & 84 \\
    Flow1D & 1.98 & 3.27 & 6.69 & 22.95 & 5.73M & 0.34GB & 79 & 1.42GB & 332 \\
    
    \bottomrule
    \end{tabular}
    \vspace{-4pt}
    \caption{Evaluation after training on FlyingChairs and FlyingThings3D datasets. Memory and inference time are measured for $448\times 1024$ and $1088 \times 1920$ resolutions on a V100 GPU, and the iteration numbers are 12 for RAFT and our method.
    }
    \label{tab:compare_cost_volume}
    \vspace{-12pt}
\end{table*}

\subsection{Analysis}

\paragraph{Ablation study.} We first analyze the effectiveness of key components in our proposed method in Tab.~\ref{tab:ablation}. When not using 1D self and cross attentions, our cost volume degrades to pure vertical and horizontal 1D searches, which apparently loses too much information \cite{ancona1995optical}. As a result, the performance drops a lot. We also note that from the detailed metrics in different motion magnitudes, our full model shows significant improvement on large motion ($s_{40+}$), demonstrating the effectiveness of the proposed method for large displacements. We also evaluate the effectiveness of self attention when computing the cross attention weights. By performing 1D self attention, 2D pixel relations are modeled more accurately, leading to better results. Meanwhile, the positional encoding used in the attention computation is helpful, as also demonstrated by previous works \cite{vaswani2017attention,carion2020end}.

We also analyze the role of each 3D cost volume plays in Tab.~\ref{tab:cost_volume_xy}. We observe that the performance
of horizontal or vertical flow is coupled with the correlation direction. Horizontal cost volume is mainly responsible
for the horizontal flow estimation, and similarly for the vertical cost volume. Concatenating these two cost volumes gives the
network necessary information for estimating both horizontal and vertical flow components.

\paragraph{Attention visualization.} To better understand how our proposed method works, we further visualize the learned 1D cross attentions. For 1D vertical attention, the learned attention map is of shape $H \times W \times H$, which includes the vertical attention weights for each pixel. we take of a row of pixels and visualize their vertical attentions in Fig.~\ref{fig:vis_attn_y}, and similarly for the horizontal attentions in Fig.~\ref{fig:vis_attn_x}. To investigate how attention maps change over time, we select target frames at different frame strides for comparison. It is observed that the learned attentions are quite sparse and generally focus on the most relevant pixels, verifying the working mechanism of our proposed method.

\begin{figure}[t]
    \centering
    \includegraphics[width=0.97\linewidth]{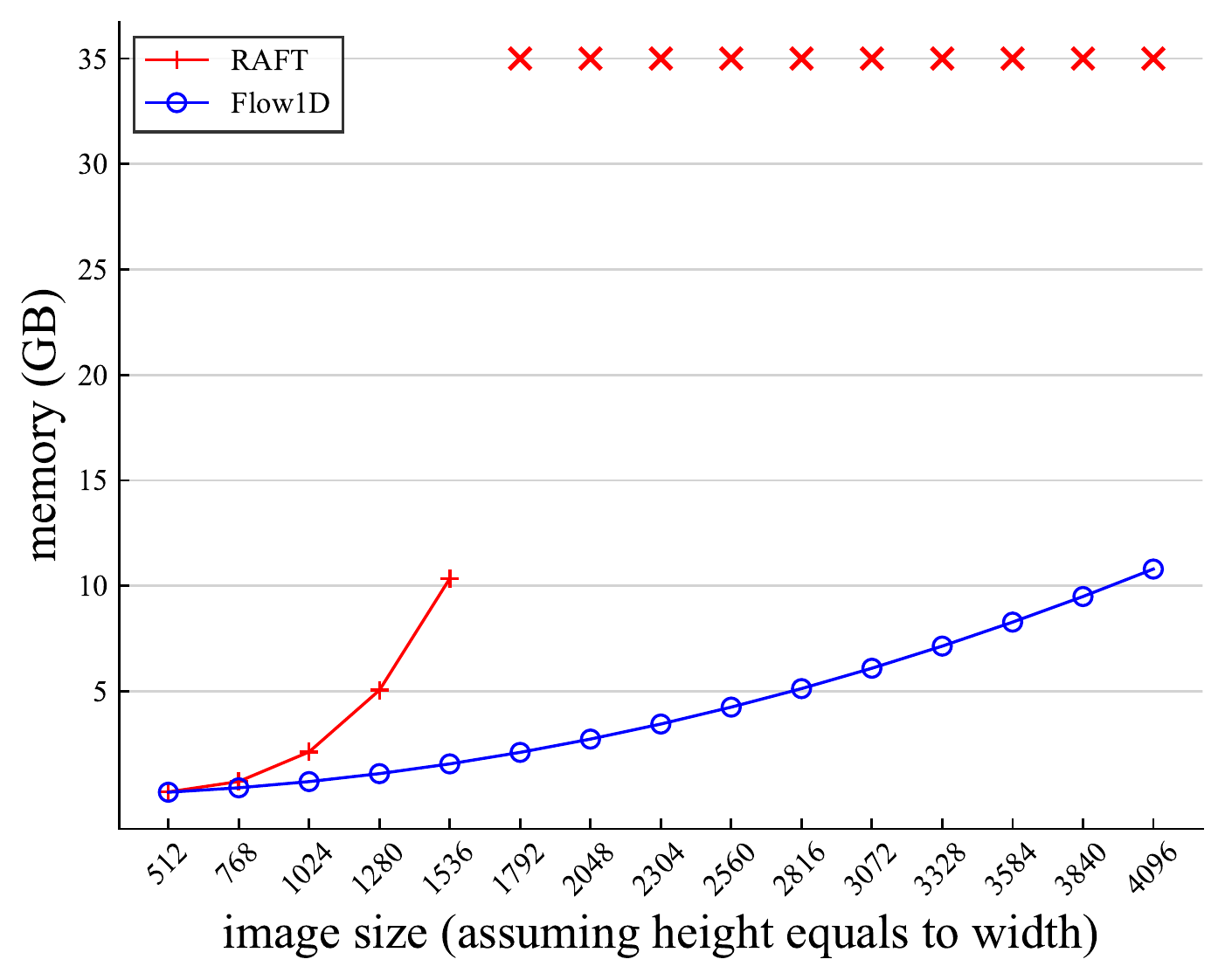}
    \vspace{-10pt}
    \caption{Memory consumption \vs. input resolutions for RAFT and our method. The red cross {\color{red}$\times$} denotes that RAFT causes an out-of-memory error for high-resolution inputs even on a 32GB GPU.}
    \label{fig:memory}
    \vspace{-10pt}
\end{figure}

\subsection{Comparison with Existing Cost Volumes}

To demonstrate the superiority of our proposed cost volume construction method, we conduct comprehensive comparisons with existing cost volumes from different aspects.

\paragraph{Setup.} We mainly compare with three representative cost volume construction methods: FlowNet2 \cite{ilg2017flownet}'s single scale cost volume, with stacked networks for refinement; PWC-Net \cite{sun2018pwc}'s multiple small cost volumes with a coarse-to-fine framework and RAFT \cite{teed2020raft}'s 4D cost volume to estimate optical flow iteratively. 
The performance is evaluated after training on FlyingChairs and FlyingThings3D datasets.

\paragraph{Sintel and KITTI results.} Table~\ref{tab:compare_cost_volume} shows the comprehensive evaluation results. In terms of accuracy, our method is higher than FlowNet2 and PWC-Net, especially on KITTI dataset, but is inferior to RAFT. 
Exhaustively constructing a large 4D cost volume is indeed advantageous to obtain highly accurate flow in RAFT, while our 3D cost volume may have the risk of missing relevant pixels. This phenomenon can be partially observed in the visualization of the learned attention maps in Fig.~\ref{fig:vis_attn_y} and Fig.~\ref{fig:vis_attn_x}: although most pixels are able to find the right correspondences, there are few noises in the attention maps that cause missing pixels. It's possible that a better design of the attention matrix computation will further improve the performance. However, our method shows higher efficiency than RAFT in terms of both memory consumption and inference speed. The superiority will be more significant for high-resolution images. For example, at $1088 \times 1920$ image resolution, our method consumes $6 \times$ less memory than RAFT\footnote{Note although RAFT can reduce the memory by re-constructing cost volume at each iteration with a customized CUDA implementation, the inference time increases substantially in practice ($\sim$ 4$\times$ slower)}.

\paragraph{High-resolution results.} We further show some visual comparisons on the high-resolution ($1080 \times 1920$) DAVIS \cite{Pont-Tuset_arXiv_2017} dataset in Fig.~\ref{fig:high_res}. We can achieve comparable results with RAFT while consuming $6 \times $ less memory. We also show additional results on 4K ($2160\times 3840$) resolution images in Fig.~\ref{fig:4k}, which can not be processed by RAFT due to the huge memory consumption. We achieve satisfactory optical flow estimation while consuming only 5.8GB memory. More results can be found in \emph{supplementary materials}.

\paragraph{Scalability.} We further compare the scalability of our method with RAFT. For input feature map of size $H \times W \times D$, the computational complexity of constructing a 4D cost volume in RAFT \cite{teed2020raft} is $\mathcal{O} ((HW)^2 D)$. As a comparison, ours is $\mathcal{O} (HW (H + W) D)$ for two 3D cost volumes. We measure the practical memory consumption under different input resolutions in Fig.~\ref{fig:memory}, our method is able to scale to more than 8K resolution ($4320 \times 7680$, memory consumption is 21.81GB) images while RAFT quickly causes an out-of-memory issue even on a high-end GPU that has 32GB memory, demonstrating the superiority of our method.

\subsection{Benchmark Results}

\begin{table}[t]
\small
    \centering
    \setlength{\tabcolsep}{3.pt} %
    
    \begin{tabular}{lcccccc}
    \toprule
    \multirow{2}{*}[-2pt]{Method } & \multicolumn{2}{c}{Sintel (train)} &
    \multicolumn{2}{c}{Sintel (test)} & \multicolumn{2}{c}{KITTI (F1-all)} \\
    \addlinespace[-12pt]  \\
    \cmidrule(lr){2-3} \cmidrule(lr){4-5} \cmidrule(lr){6-7}
    \addlinespace[-12pt] \\
    & Clean & Final & Clean & Final & (train) & (test) \\
    \midrule
    
    FlowNet2 \cite{ilg2017flownet} & (1.45) & (2.01) & 4.16 & 5.74 & (6.8) & 11.48 \\
    PWC-Net+ \cite{sun2019models} & (1.71) & (2.34) & 3.45 & 4.60 & (5.3) & 7.72 \\
    HD$^3$ \cite{yin2019hierarchical} & (1.87) & (1.17) & 4.79 & 4.67 & (4.1) & 6.55 \\
    VCN \cite{yang2019volumetric} & (1.66) & (2.24) & 2.81 & 4.40 & (4.1) & 6.30 \\
    MaskFlowNet \cite{zhao2020maskflownet} & - & - & 2.52 & 4.17 & - & 6.10 \\
    RAFT \cite{teed2020raft} & (0.77) & (1.27) & 1.61 & 2.86 & (1.5) & 5.10 \\
    Flow1D & (0.84) & (1.25) & 2.24 & 3.81 & (1.6) & 6.27 \\
    
    \bottomrule
    \end{tabular}
    \caption{Benchmark performance on Sintel and KITTI datasets. The numbers in the parenthesis are the results on the data that the methods have been fine-tuned on.}
    \label{tab:sintel_kitti}
\end{table}

\paragraph{Sintel.} For submission to Sintel dataset, we fine-tune on mixed KITTI \cite{geiger2012we,menze2015object}, HD1K \cite{kondermann2016hci}, FlyingThings3D \cite{mayer2016large} and Sintel \cite{butler2012naturalistic} datasets for 100K iterations. The evaluation results are shown in Tab.~\ref{tab:sintel_kitti}. Our method ranks second only to RAFT, outperforming previous representative methods such as PWC-Net \cite{sun2018pwc} and FlowNet2 \cite{ilg2017flownet}.

\paragraph{KITTI.} We further fine-tune on the KITTI 2015 training set for 50K iterations. Table~\ref{tab:sintel_kitti} shows the evaluation results. Our method performs better than PWC-Net+, while slightly inferior to MaskFlowNet, which is likely caused by the limited training data of KITTI dataset.

\begin{figure}[t]
\centering
\setlength{\tabcolsep}{0.5pt}

{\renewcommand{\arraystretch}{0.3} %

\begin{tabular}{lcc}

\rotatebox{90}{{\quad  Image}} &
\includegraphics[width=0.48\linewidth]{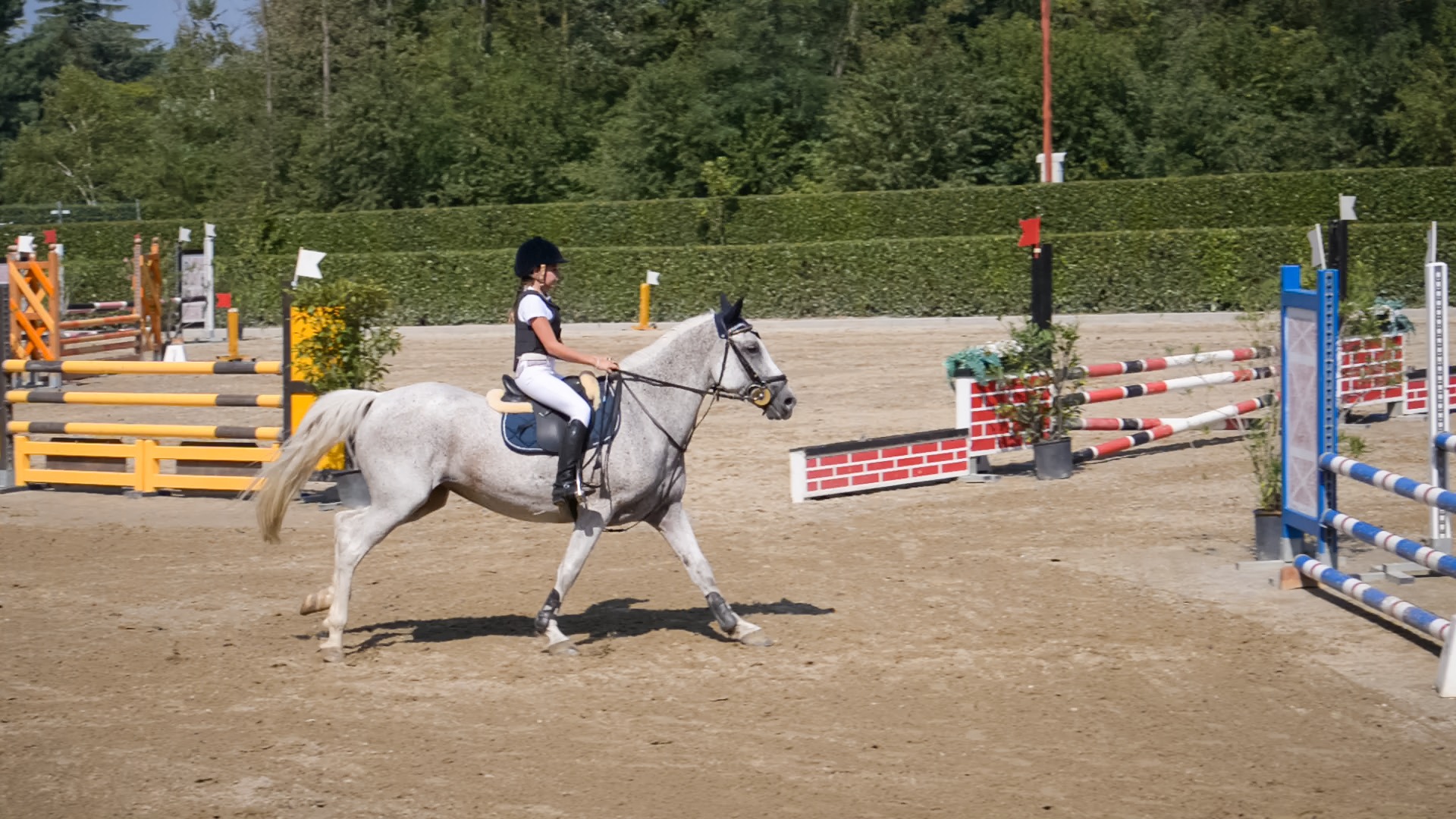} &
\includegraphics[width=0.48\linewidth]{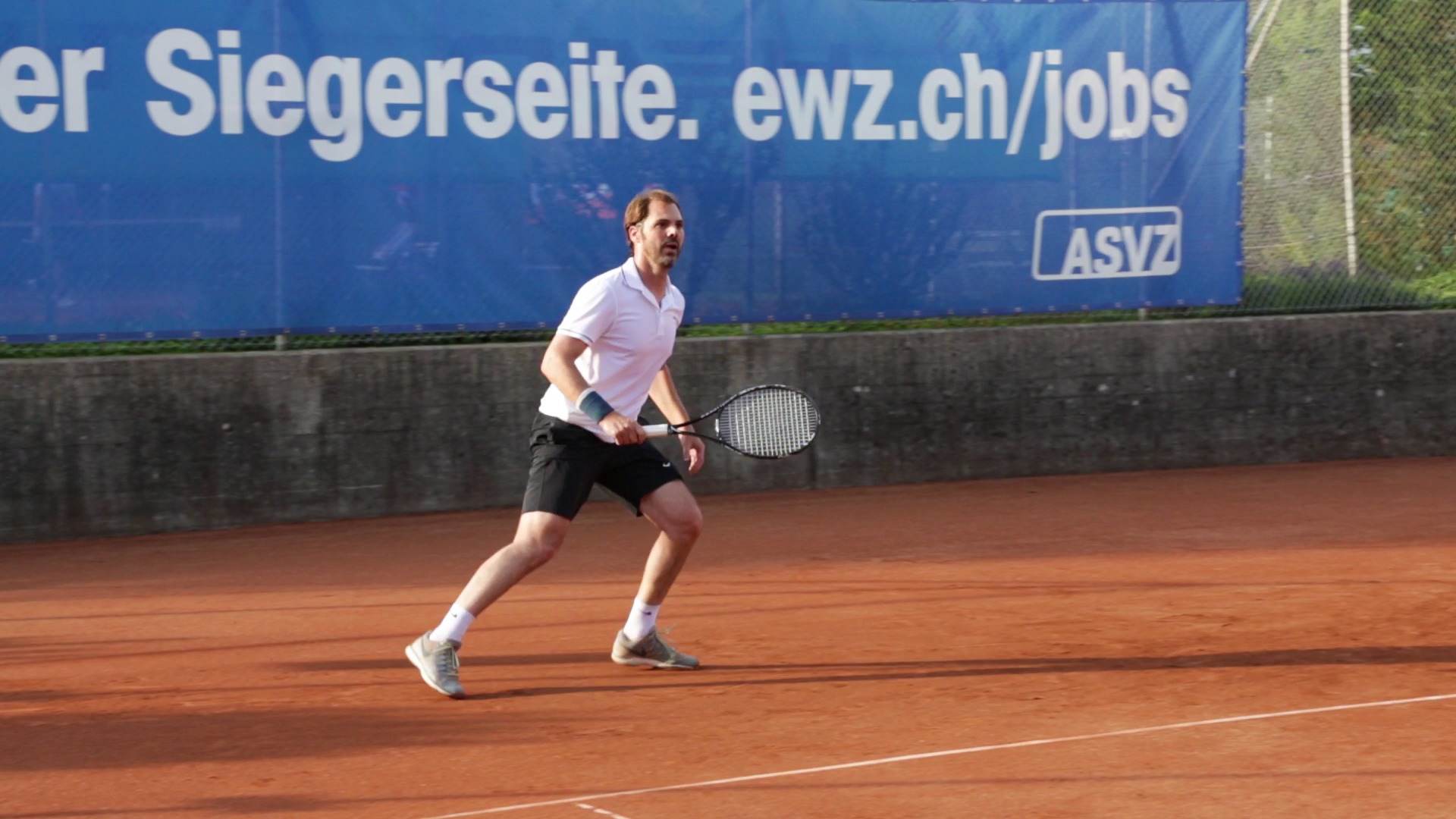} \\

\rotatebox{90}{{ \quad  RAFT }} &
\includegraphics[width=0.48\linewidth]{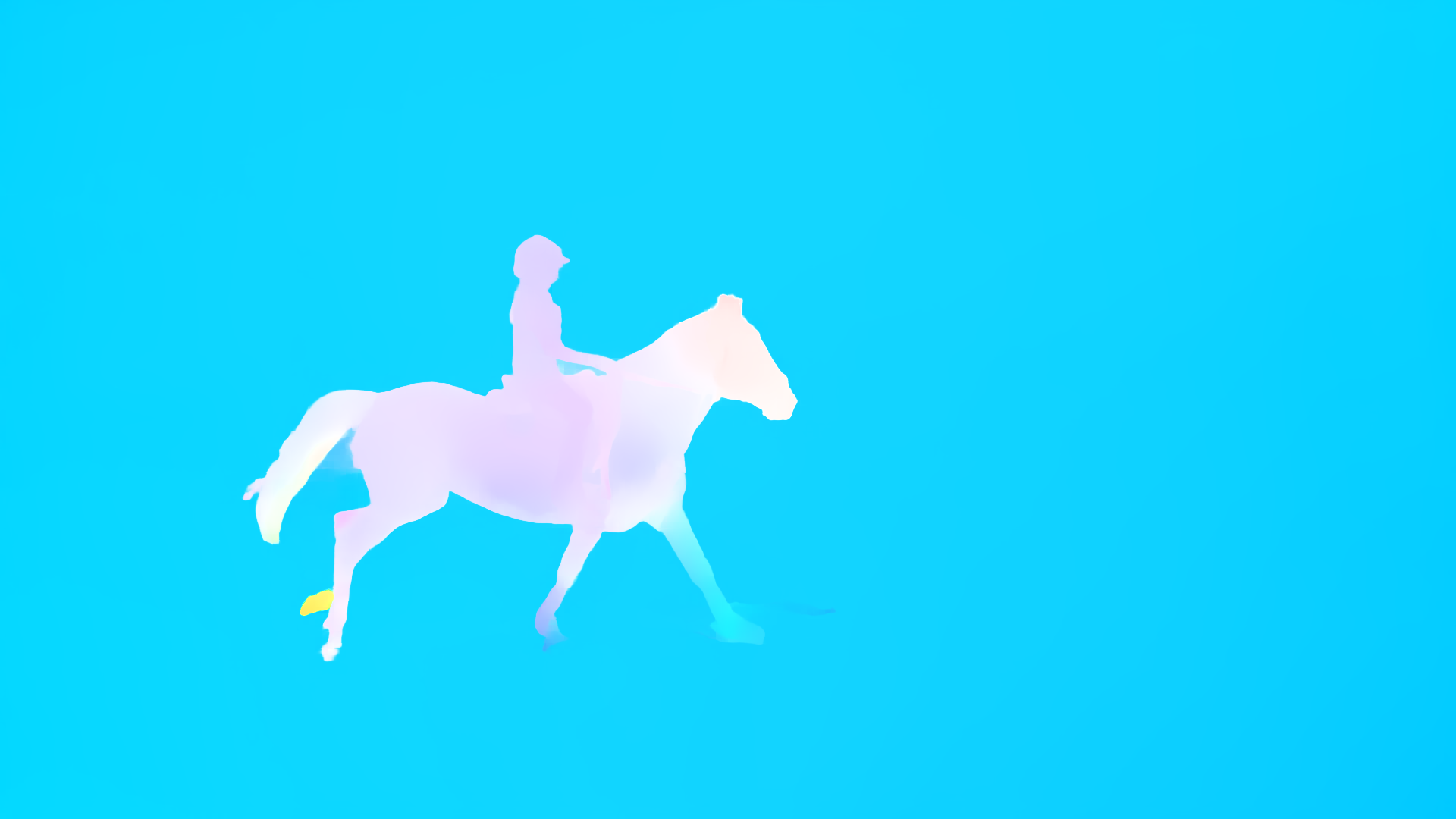} &
\includegraphics[width=0.48\linewidth]{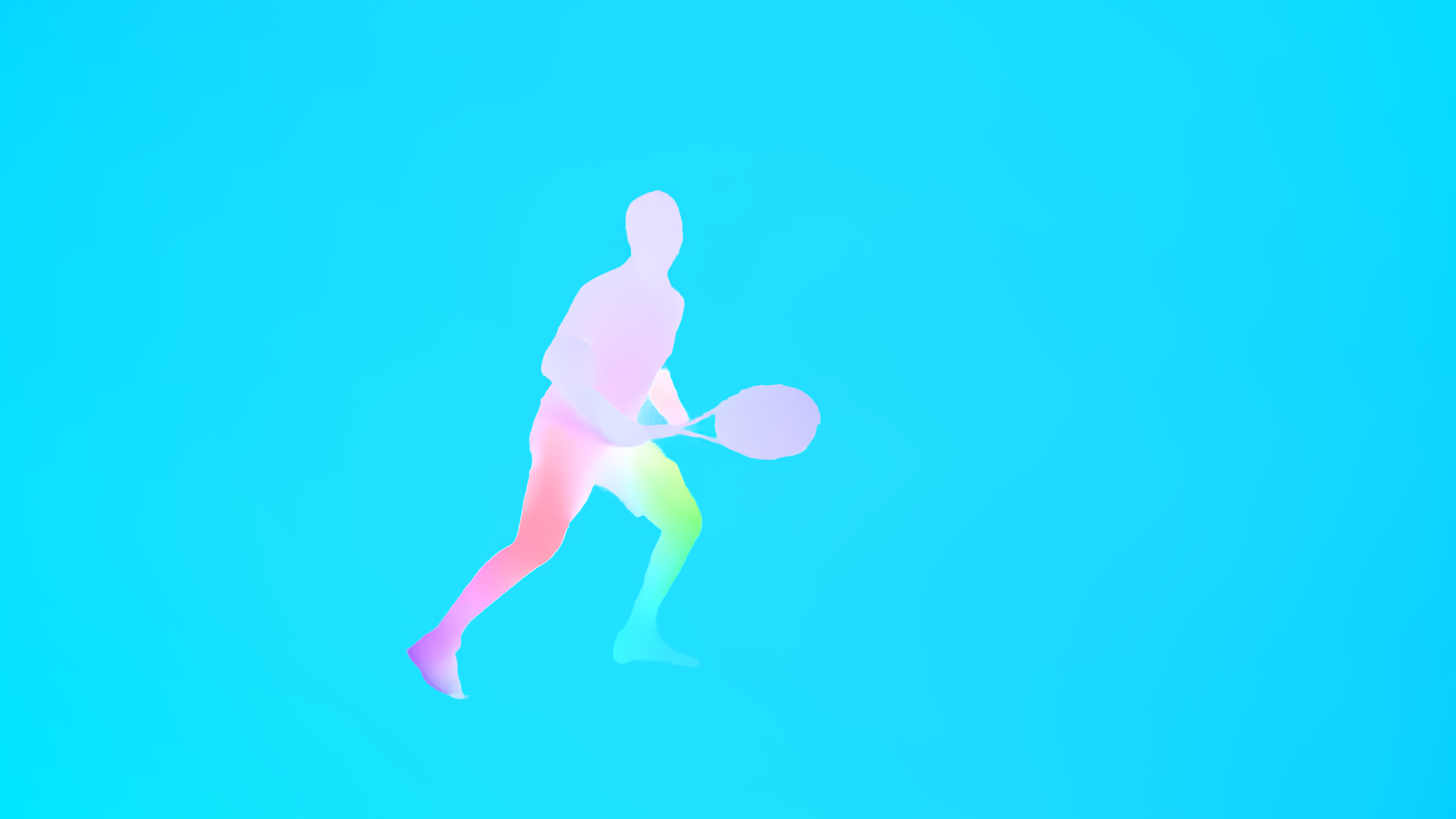} \\

\rotatebox{90}{{ \quad Flow1D}} &
\includegraphics[width=0.48\linewidth]{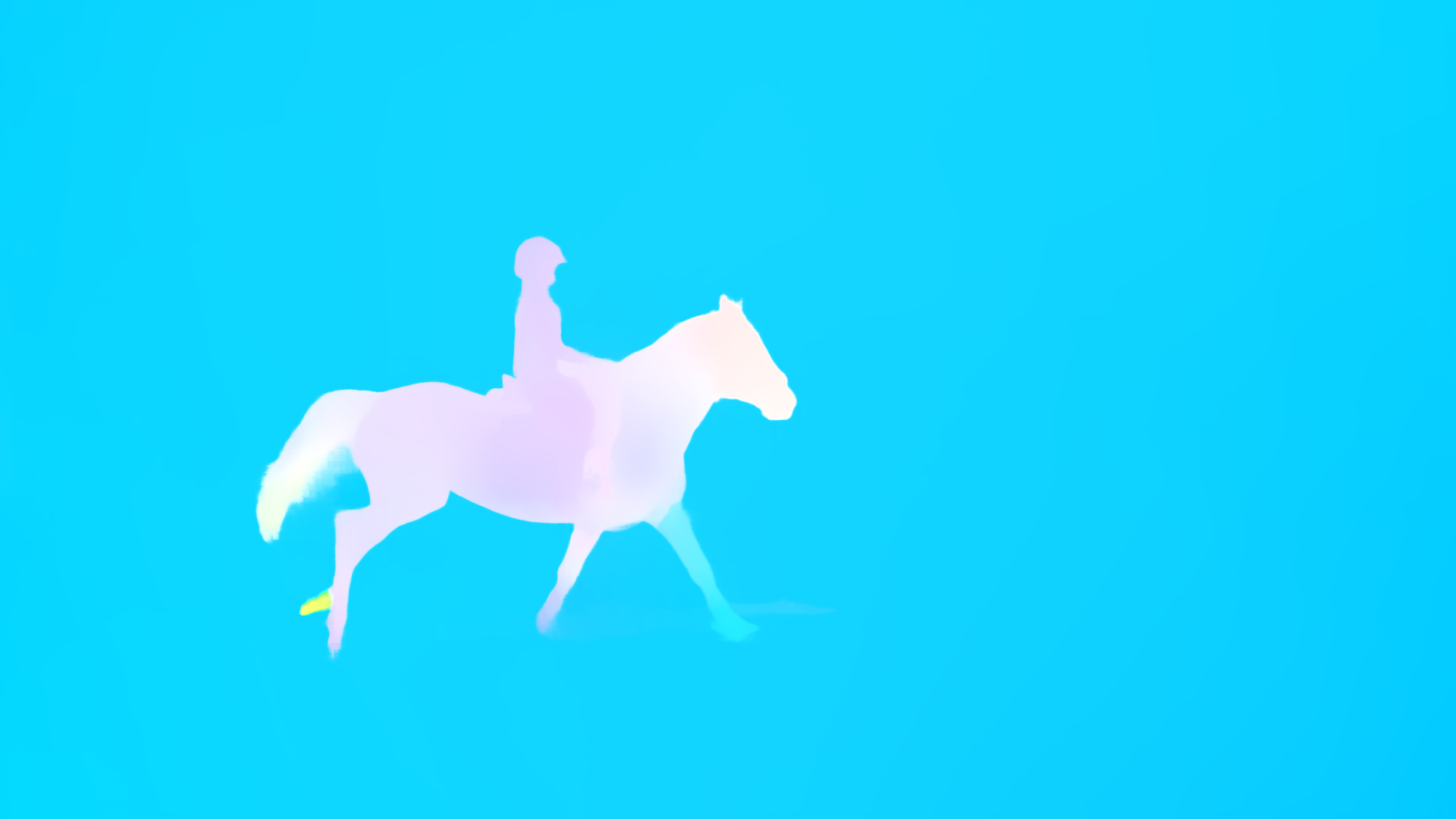} &
\includegraphics[width=0.48\linewidth]{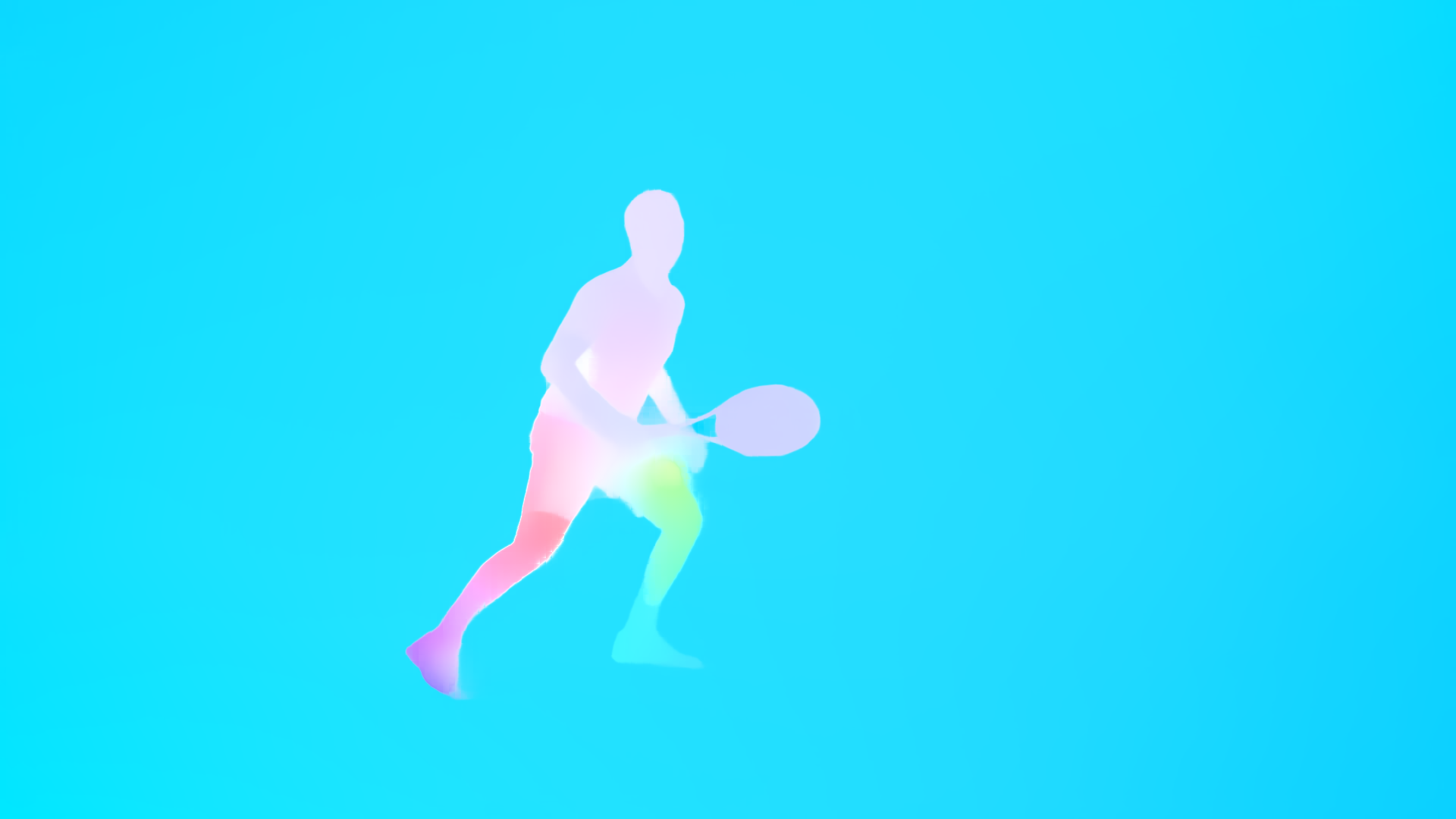} \\

\end{tabular}
}
\vspace{-7pt}
\caption{Comparisons on high-resolution ($1080 \times 1920$) images from DAVIS dataset. We achieve comparable results with RAFT while consuming $6 \times$ less memory.}
\label{fig:high_res}
\end{figure}

\begin{figure}[t]
\centering
\vspace{-8pt}
\setlength{\tabcolsep}{0.5pt}

{\renewcommand{\arraystretch}{0.3} %

\begin{tabular}{cc}

\includegraphics[width=\linewidth]{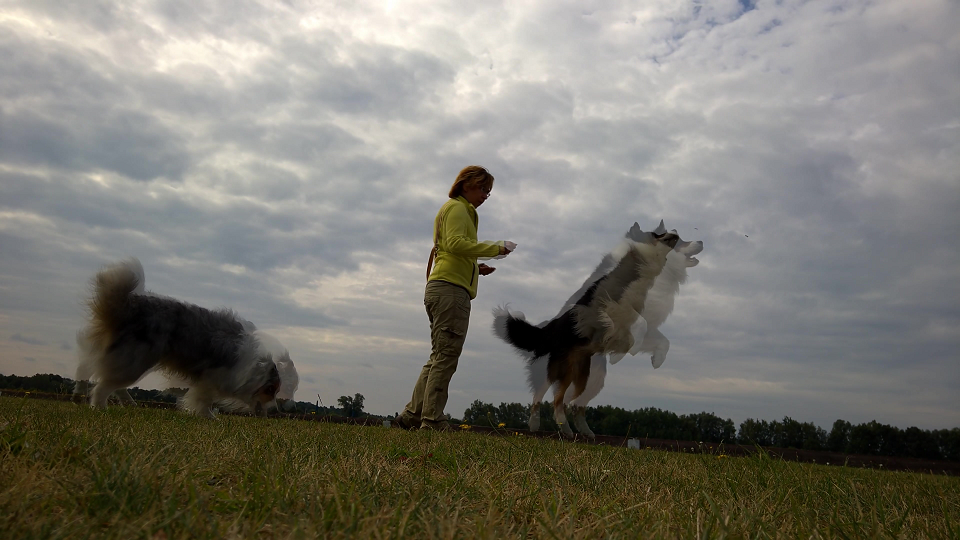} \\
\includegraphics[width=\linewidth]{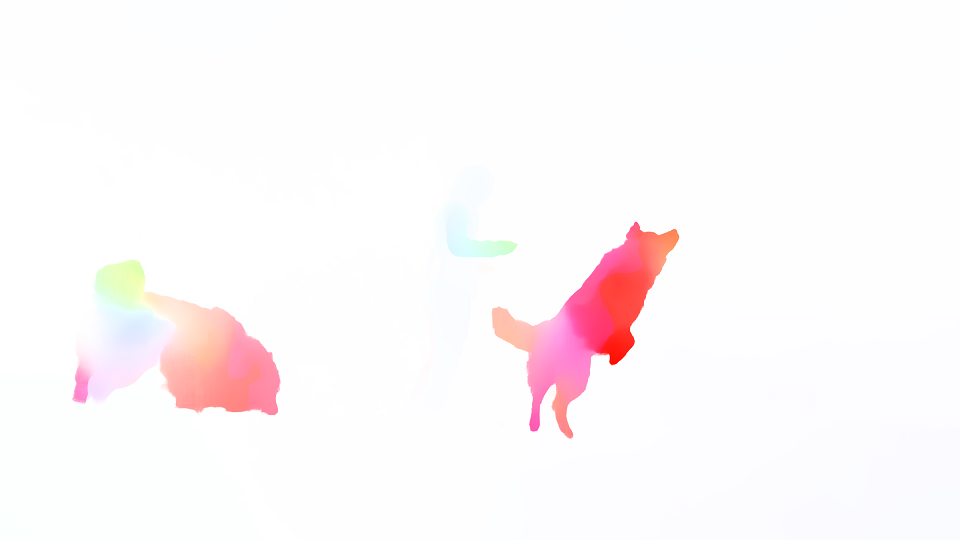} \\

\end{tabular}
}
\caption{Optical flow prediction results on 4K resolution ($2160 \times 3840$) images from DAVIS dataset.}
\label{fig:4k}
\vspace{-8pt}
\end{figure}

\section{Conclusion}

We have presented a new cost volume construction method for high-resolution optical flow estimation. By factorizing the 2D optical flow with 1D attention and 1D correlation, we are able to scale to more than 8K resolution images while maintaining competitive performance. 
We hope our new perspective can stimulate future research on cost volume compression and efficient high-resolution optical flow estimation.

\small {\noindent{\bf Acknowledgement.} This work was partially supported by NSFC (No.~62122071), the Youth Innovation Promotion Association CAS (No.~2018495) and the Fundamental Research Funds for the Central Universities (No.~WK3470000021).}

{\small
\bibliographystyle{ieee_fullname}
\bibliography{egbib}
}

\newpage
\onecolumn

\section*{Appendix}
\renewcommand{\thesubsection}{\Alph{subsection}}

In this supplementary document, we first present additional evaluations of our proposed method. Then we provide more visual results on 4K ($2160 \times 3840$) resolution images from DAVIS dataset and real-world scenes captured by a mobile phone. Finally, we present additional visual results on Sintel test set and more implementation details.

\subsection{Additional Evaluations}

In the main paper, we have analyzed the role of each 3D cost volume plays and the evaluation results on Sintel (train, clean) shows that the performance of horizontal or vertical flow is coupled with the correlation direction. Here we present additional evaluations on Sintel (train, final) and KITTI (train) datasets and observe consistent results: horizontal cost volume is mainly responsible for the horizontal flow estimation, and similarly for the vertical cost volume. Concatenating these two cost volumes gives the network necessary information for estimating both horizontal and vertical flow components.

\begin{table*}[h]
	\centering
	
	\begin{tabular}{lccccccccc}
		\toprule
		\multirow{2}{*}[-2pt]{Cost volume} & \multicolumn{3}{c}{Sintel (train, clean)} & \multicolumn{3}{c}{Sintel (train, final)} & \multicolumn{3}{c}{KITTI (train)} \\
		\addlinespace[-12pt] \\
		\cmidrule(lr){2-4} \cmidrule(lr){5-7} \cmidrule(lr){8-10}
		\addlinespace[-12pt] \\
		& EPE & EPE ($x$) & EPE ($y$) & EPE & EPE ($x$) & EPE ($y$) & EPE & EPE ($x$) & EPE ($y$) \\
		\midrule
		$y$ attn, $x$ corr & 3.10 & 1.66 & 2.12 & 4.59 & 2.76 & 2.92 & 10.39 & 7.87 & 5.15 \\
		$x$ attn, $y$ corr & 4.05 & 3.55 & 1.13 & 5.66 & 4.75 & 2.03 & 14.37 & 13.71 & 2.84 \\
		concat both & {\bf 1.98} & {\bf 1.48} & {\bf 0.94} & {\bf 3.27} & {\bf 2.35} & {\bf 1.73} & {\bf 6.69} & {\bf 6.00} & {\bf 2.16} \\

		\bottomrule
	\end{tabular}
	\caption{Analysis on horizontal ($x$) and vertical ($y$) cost volumes. EPE ($x$) and EPE ($y$) represent the end-point-error of the horizontal and vertical flow component, respectively.
	}
	\label{tab:cost_volume_xy}
\end{table*}

\subsection{More Results on 4K Resolution}

\begin{figure*}[t]
	\centering
	\vspace{-8pt}
	\setlength{\tabcolsep}{0.5pt}
	
	{\renewcommand{\arraystretch}{0.3} %
		
		\begin{tabular}{cc}

			\includegraphics[width=\linewidth]{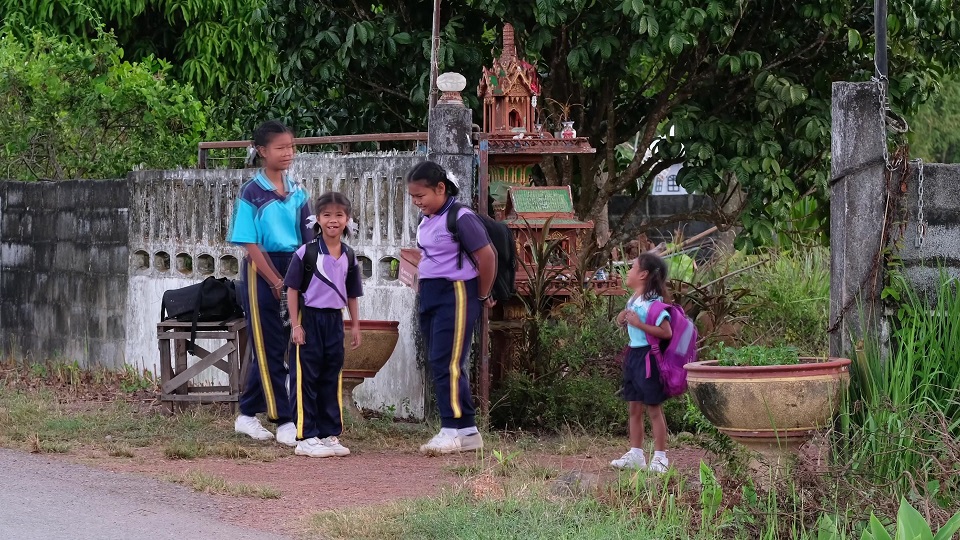} \\
			\includegraphics[width=\linewidth]{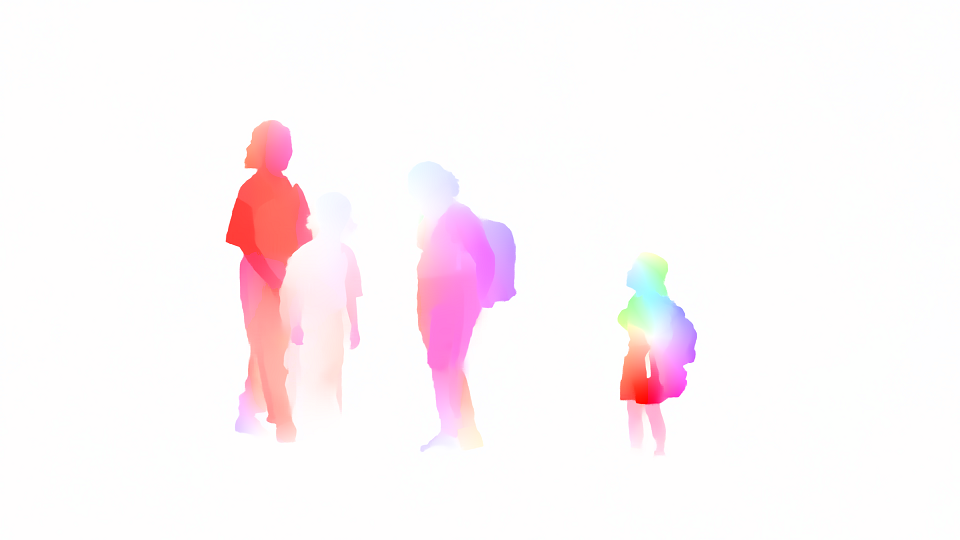} \\

		\end{tabular}
	}
	\caption{Optical flow prediction results on 4K resolution ($2160 \times 3840$) images from DAVIS dataset.}
	\label{fig:davis1}
	
\end{figure*}

\begin{figure*}[t]
	\centering
	\vspace{-8pt}
	\setlength{\tabcolsep}{0.5pt}
	
	{\renewcommand{\arraystretch}{0.3} %
		
		\begin{tabular}{cc}

			\includegraphics[width=\linewidth]{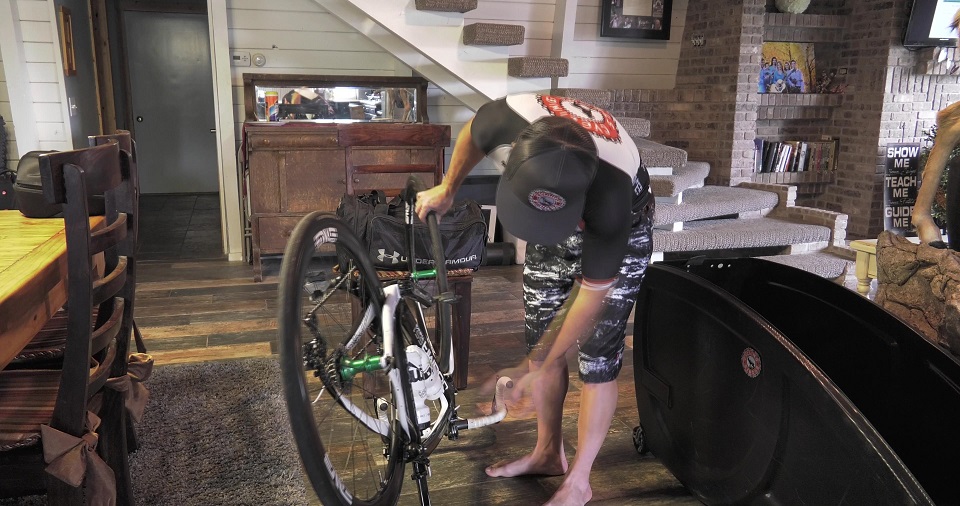} \\
			\includegraphics[width=\linewidth]{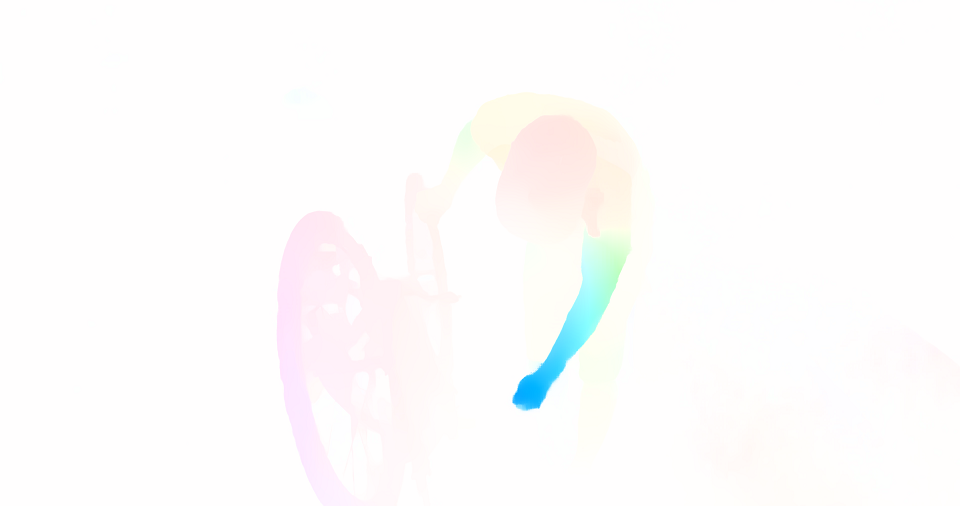} \\

		\end{tabular}
	}
	\caption{Optical flow prediction results on 4K resolution ($2160 \times 3840$) images from DAVIS dataset.}
	\label{fig:davis2}
	
\end{figure*}

\begin{figure*}[t]
	\centering
	\vspace{-8pt}
	\setlength{\tabcolsep}{0.5pt}
	
	{\renewcommand{\arraystretch}{0.3} %
		
		\begin{tabular}{cc}

			\includegraphics[width=\linewidth]{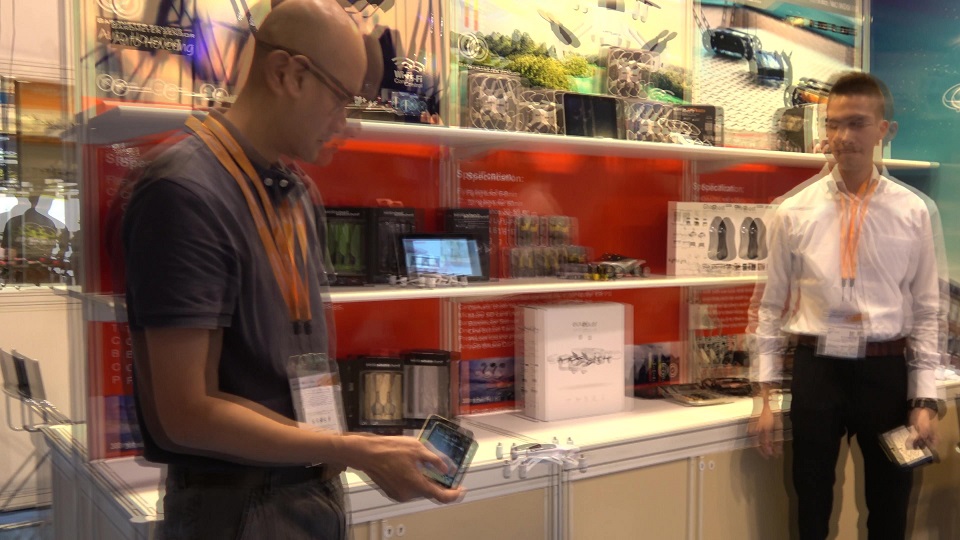} \\
			\includegraphics[width=\linewidth]{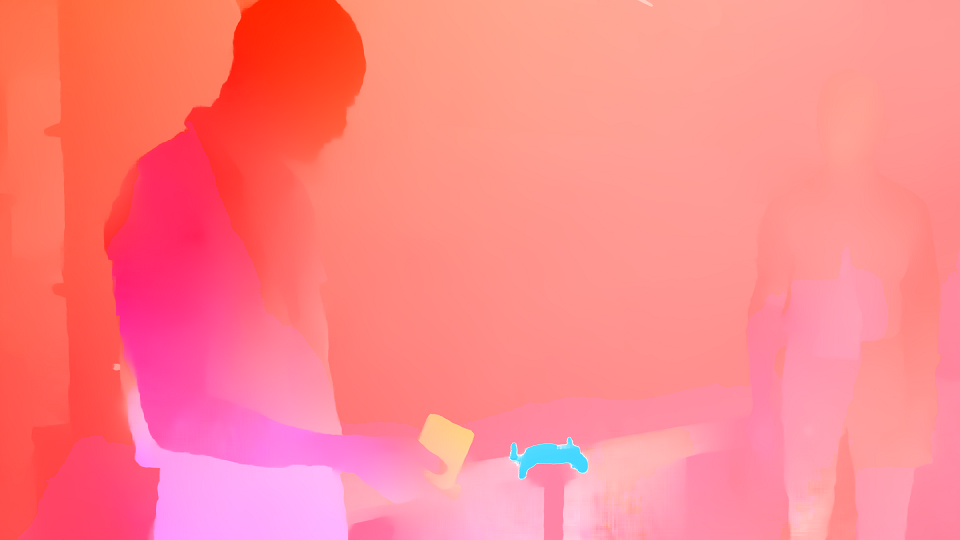} \\

		\end{tabular}
	}
	\caption{Optical flow prediction results on 4K resolution ($2160 \times 3840$) images from DAVIS dataset.}
	\label{fig:davis3}
	
\end{figure*}

\begin{figure*}[t]
	\centering
	\vspace{-8pt}
	\setlength{\tabcolsep}{0.5pt}
	
	{\renewcommand{\arraystretch}{0.3} %
		
		\begin{tabular}{cc}

			\includegraphics[width=\linewidth]{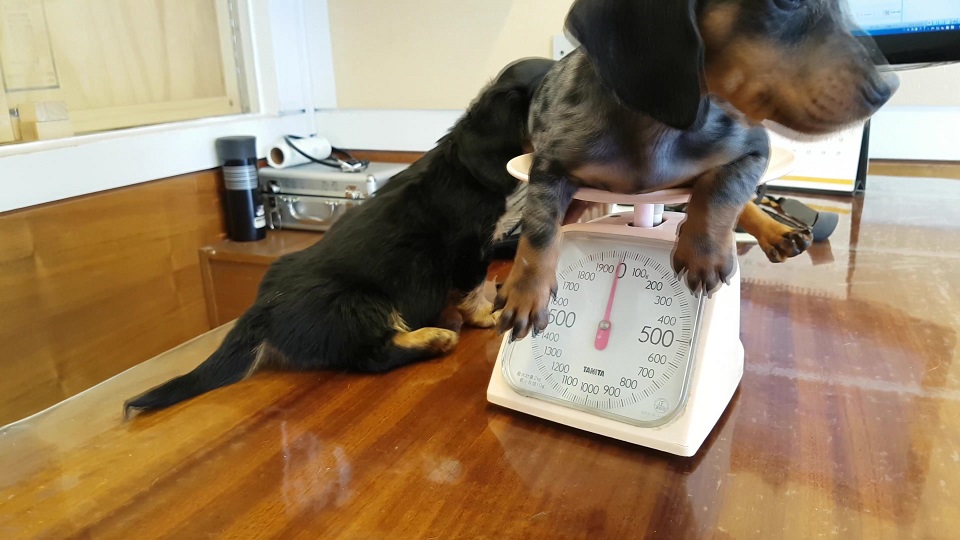} \\
			\includegraphics[width=\linewidth]{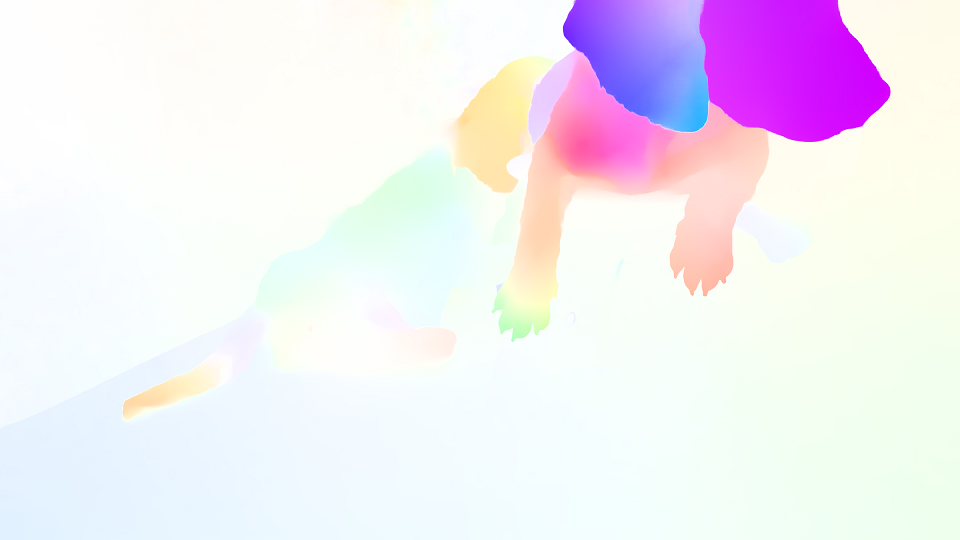} \\

		\end{tabular}
	}
	\caption{Optical flow prediction results on 4K resolution ($2160 \times 3840$) images from DAVIS dataset.}
	\label{fig:davis4}
	
\end{figure*}

\begin{figure*}[t]
	\centering
	\vspace{-8pt}
	\setlength{\tabcolsep}{0.5pt}
	
	{\renewcommand{\arraystretch}{0.3} %
		
		\begin{tabular}{cc}

			\includegraphics[width=\linewidth]{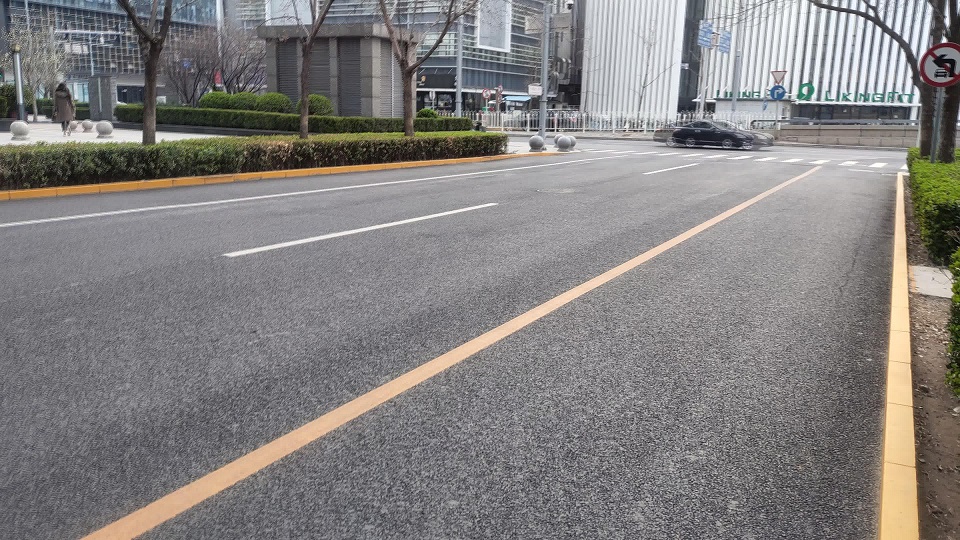} \\
			\includegraphics[width=\linewidth]{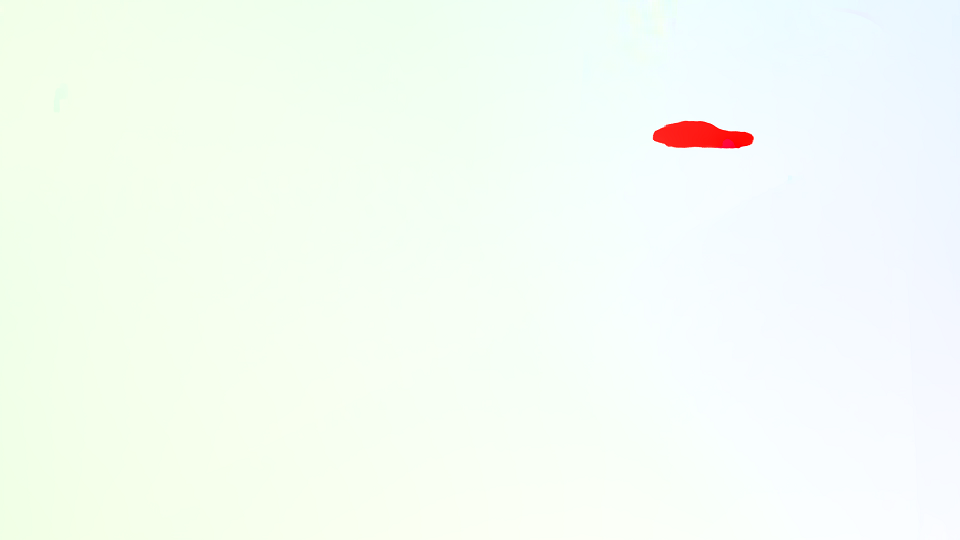} \\

		\end{tabular}
	}
	\caption{Optical flow prediction results on real-world 4K resolution ($2160 \times 3840$) images captured by a mobile phone.}
	\label{fig:phone1}
	
\end{figure*}

\begin{figure*}[t]
	\centering
	\vspace{-8pt}
	\setlength{\tabcolsep}{0.5pt}
	
	{\renewcommand{\arraystretch}{0.3} %
		
		\begin{tabular}{cc}

			\includegraphics[width=\linewidth]{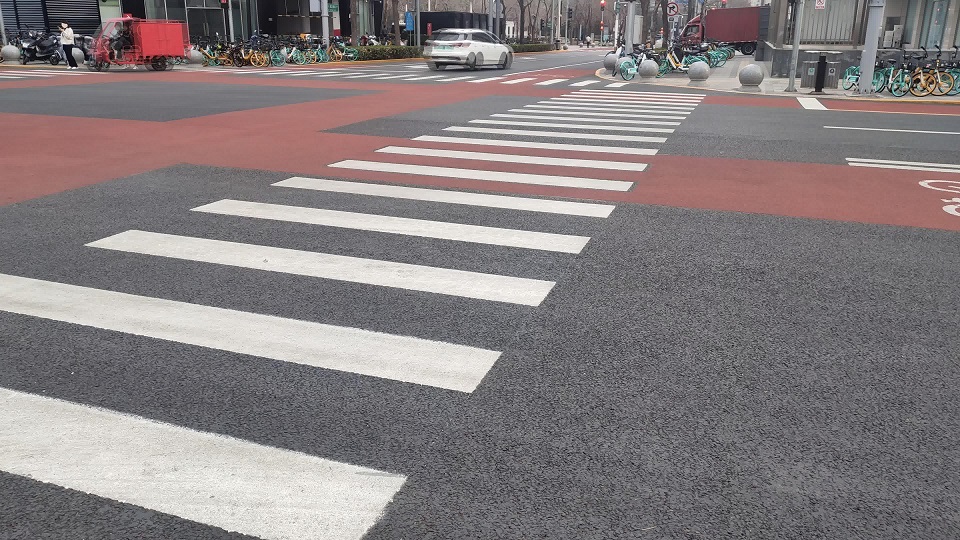} \\
			\includegraphics[width=\linewidth]{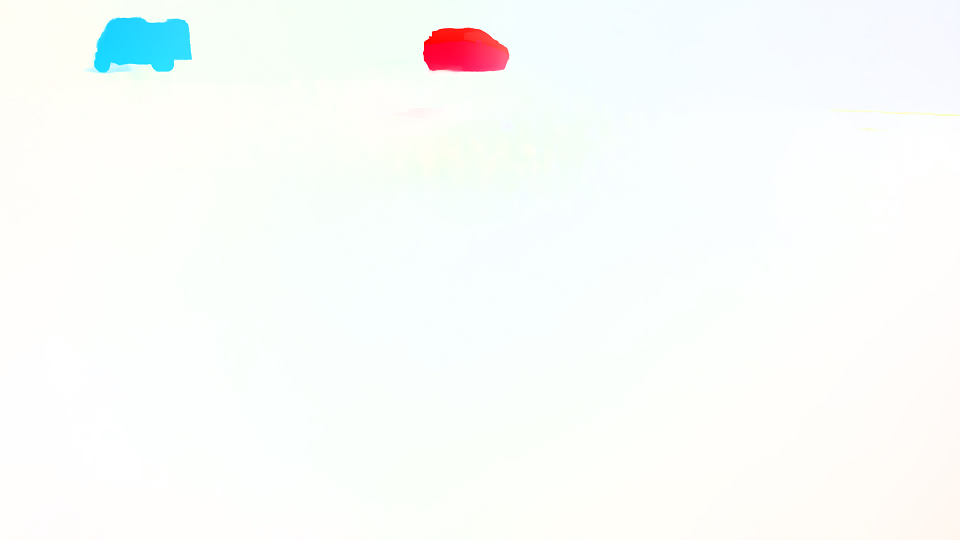} \\

		\end{tabular}
	}
	\caption{Optical flow prediction results on real-world 4K resolution ($2160 \times 3840$) images captured by a mobile phone.}
	\label{fig:phone2}
	
\end{figure*}

\begin{figure*}[t]
	\centering
	\vspace{-8pt}
	\setlength{\tabcolsep}{0.5pt}
	
	{\renewcommand{\arraystretch}{0.3} %
		
		\begin{tabular}{cc}

			\includegraphics[width=\linewidth]{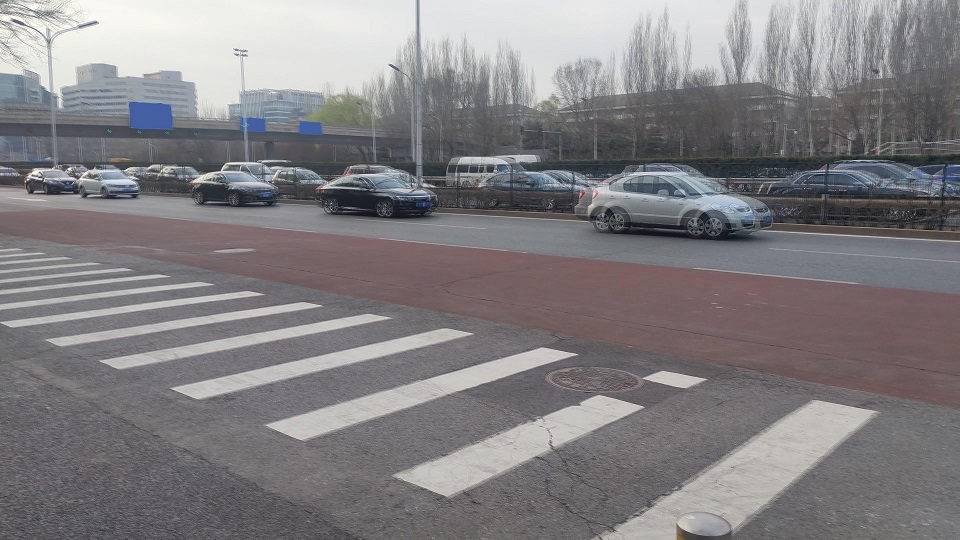} \\
			\includegraphics[width=\linewidth]{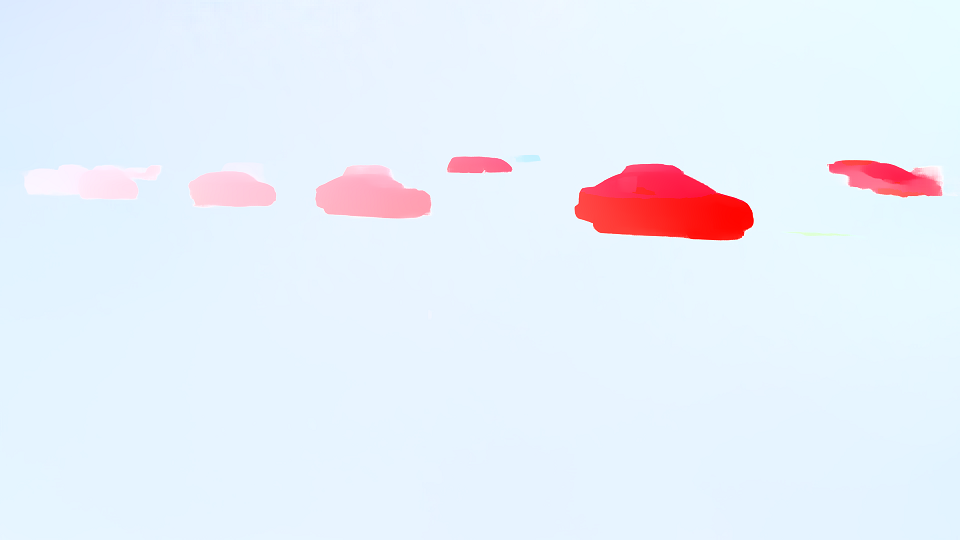} \\

		\end{tabular}
	}
	\caption{Optical flow prediction results on real-world 4K resolution ($2160 \times 3840$) images captured by a mobile phone.}
	\label{fig:phone3}
	
\end{figure*}

\begin{figure*}[t]
	\centering
	\vspace{-8pt}
	\setlength{\tabcolsep}{0.5pt}
	
	{\renewcommand{\arraystretch}{0.3} %
		
		\begin{tabular}{cc}

			\includegraphics[width=\linewidth]{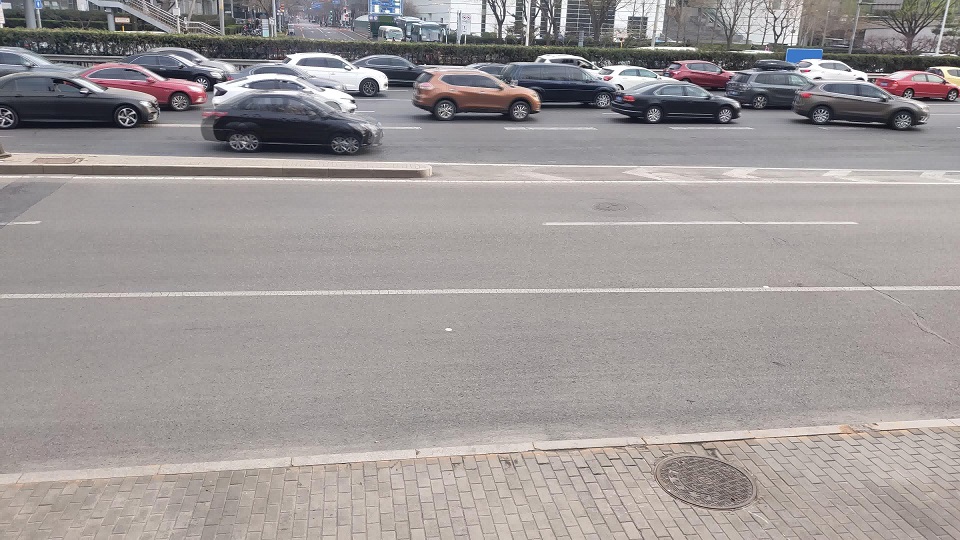} \\
			\includegraphics[width=\linewidth]{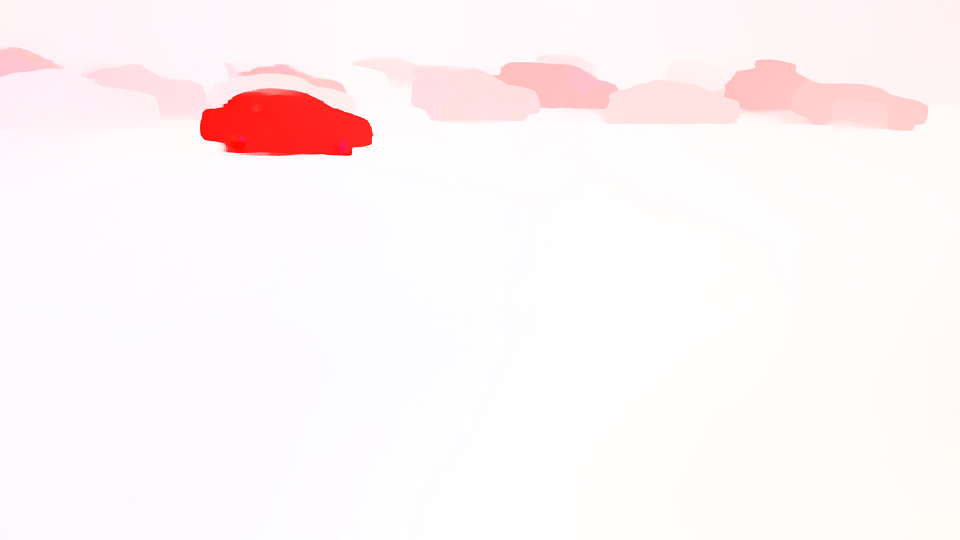} \\

		\end{tabular}
	}
	\caption{Optical flow prediction results on real-world 4K resolution ($2160 \times 3840$) images captured by a mobile phone.}
	\label{fig:phone4}
	
\end{figure*}

We provide additional visual results on 4K resolution ($2160 \times 3840$) images from DAVIS dataset in Fig.~\ref{fig:davis1}, \ref{fig:davis2}, \ref{fig:davis3}, \ref{fig:davis4}, and real-world scenes captured by a mobile phone in Fig.~\ref{fig:phone1}, \ref{fig:phone2}, \ref{fig:phone3}, \ref{fig:phone4}.

\subsection{Visual Results on Sintel}

\begin{figure*}[t]
	\centering
	\setlength{\tabcolsep}{0.5pt}
	
	{\renewcommand{\arraystretch}{0.3} %
		
		\begin{tabular}{ccccc}
			
			\includegraphics[width=0.19\linewidth]{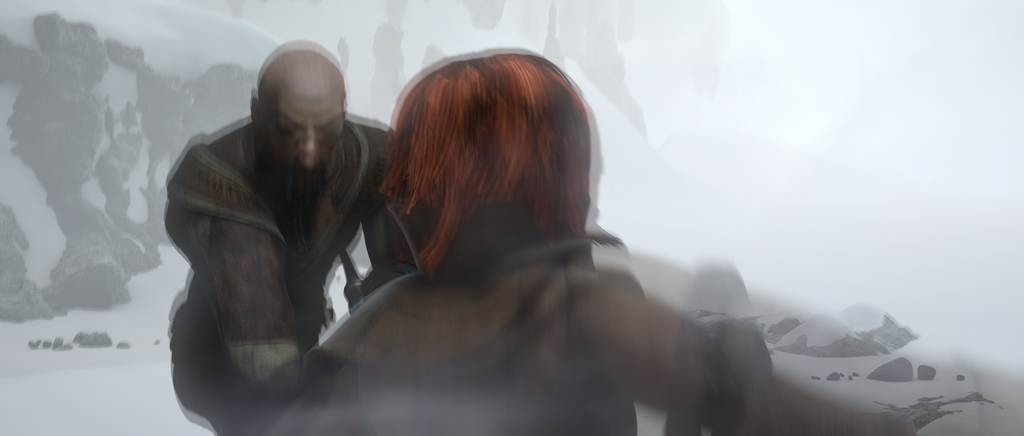} &
			\includegraphics[width=0.19\linewidth]{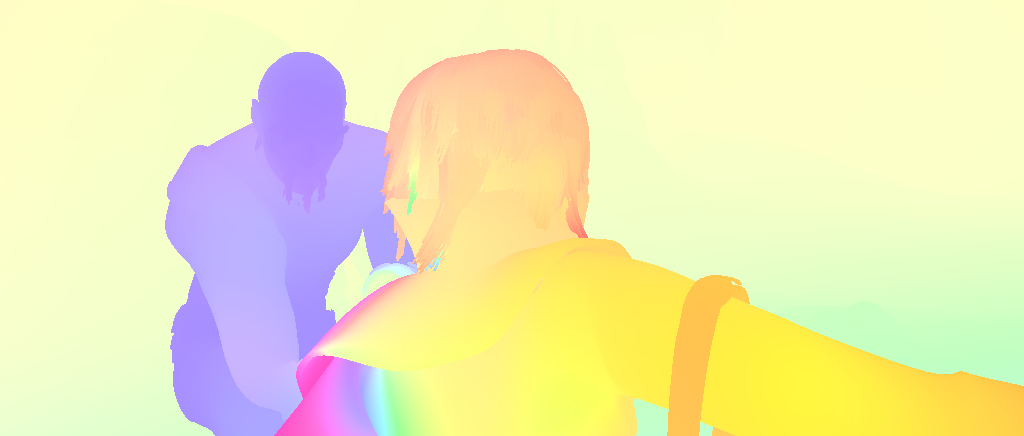} & 
			\includegraphics[width=0.19\linewidth]{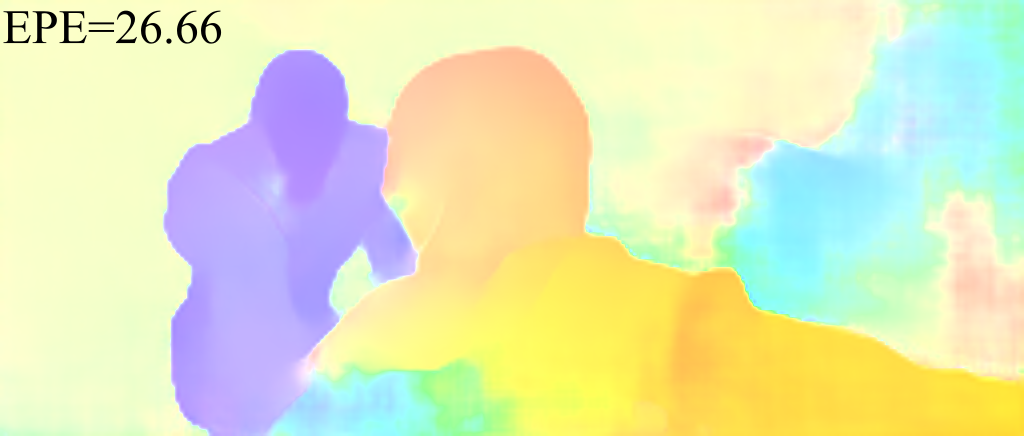} &
			\includegraphics[width=0.19\linewidth]{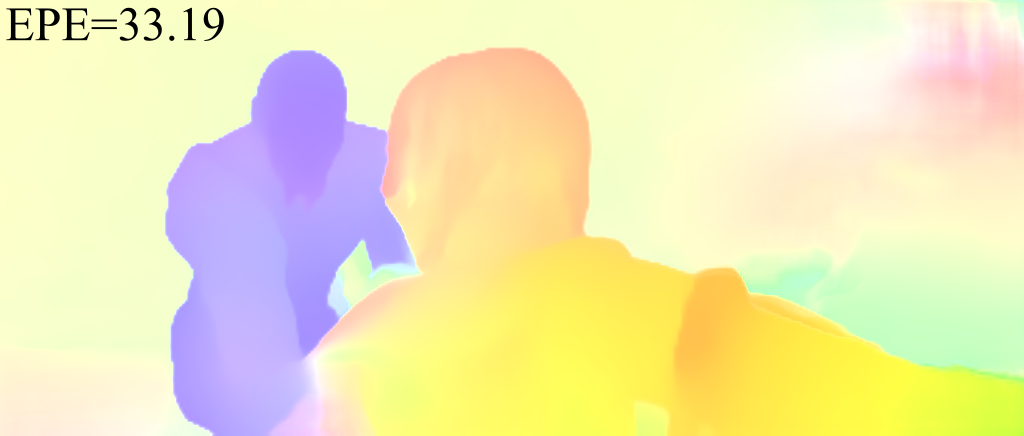} & 
			\includegraphics[width=0.19\linewidth]{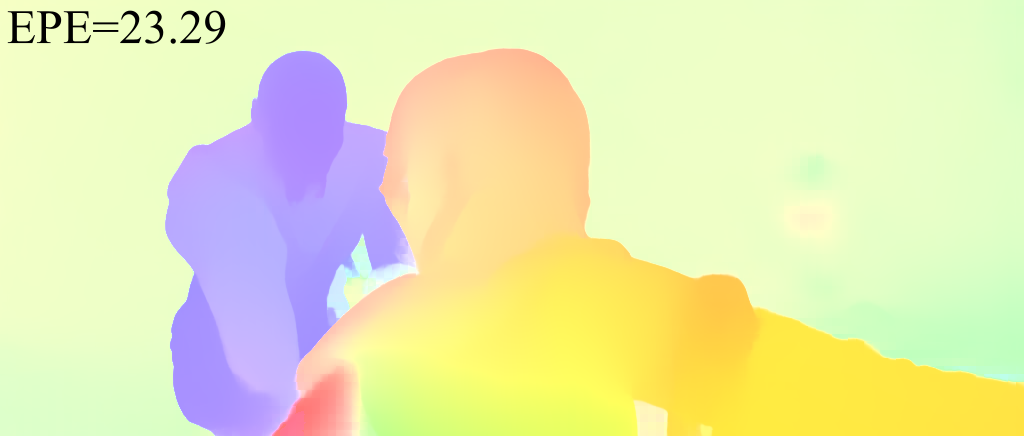} \\

			\includegraphics[width=0.19\linewidth]{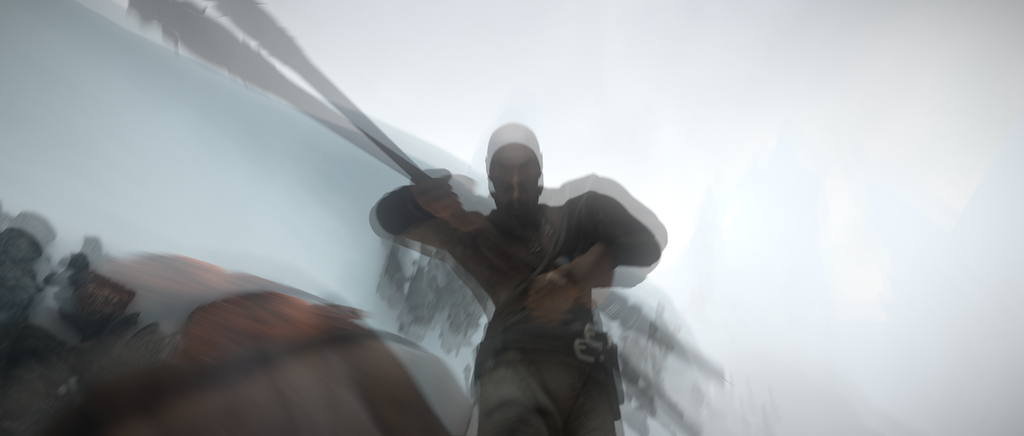} &
			\includegraphics[width=0.19\linewidth]{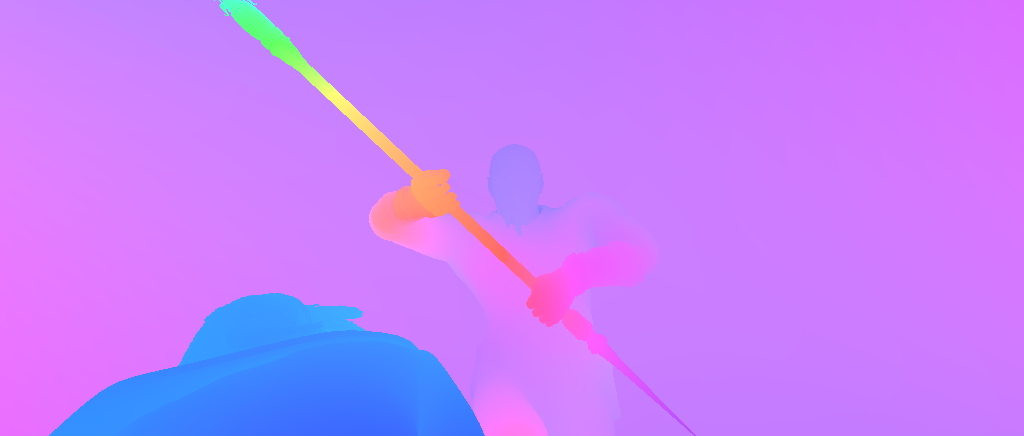} & 
			\includegraphics[width=0.19\linewidth]{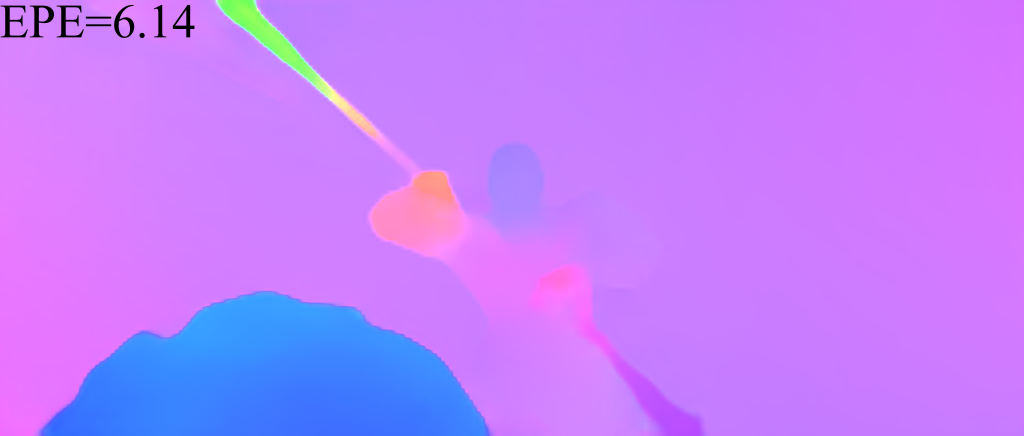} &
			\includegraphics[width=0.19\linewidth]{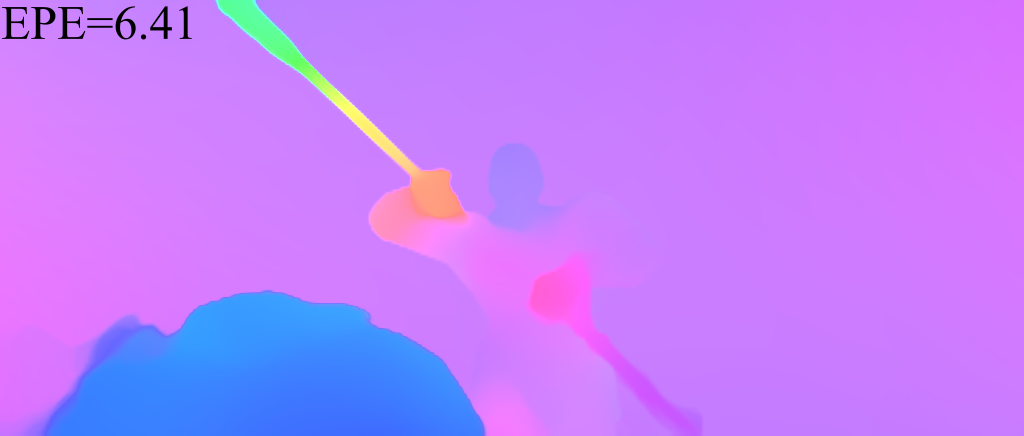} & 
			\includegraphics[width=0.19\linewidth]{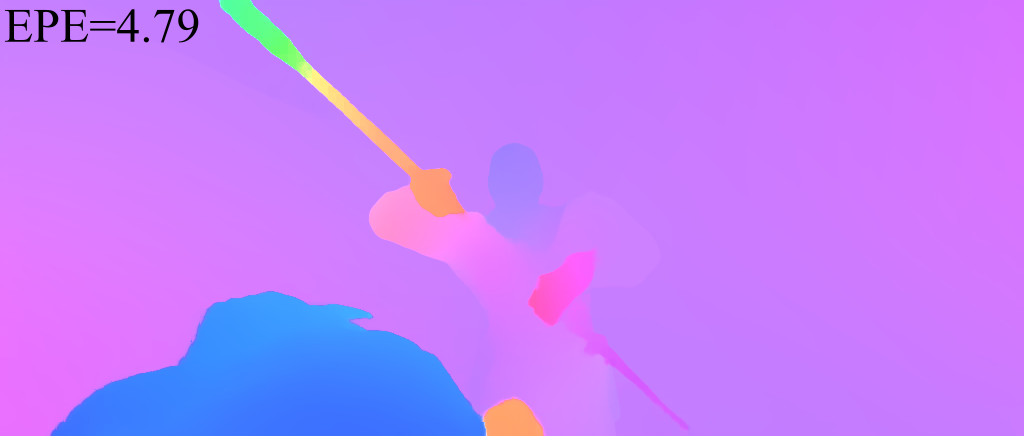} \\
			
			\includegraphics[width=0.19\linewidth]{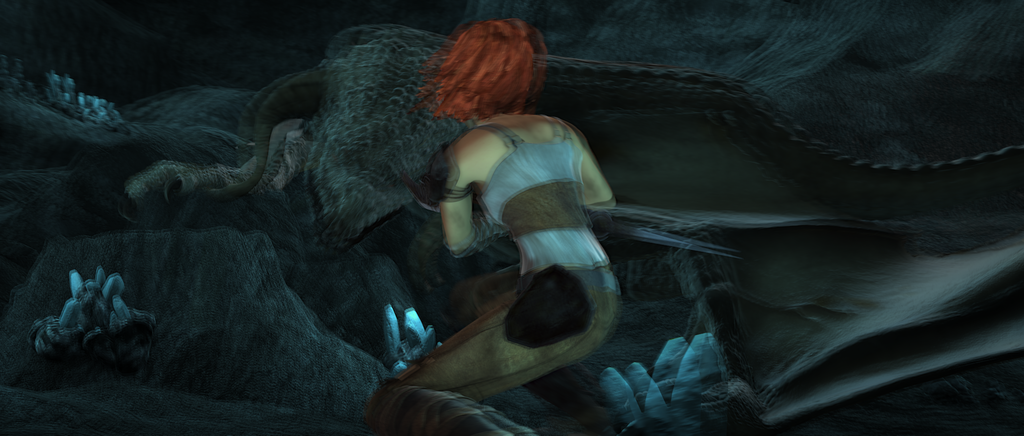} &
			\includegraphics[width=0.19\linewidth]{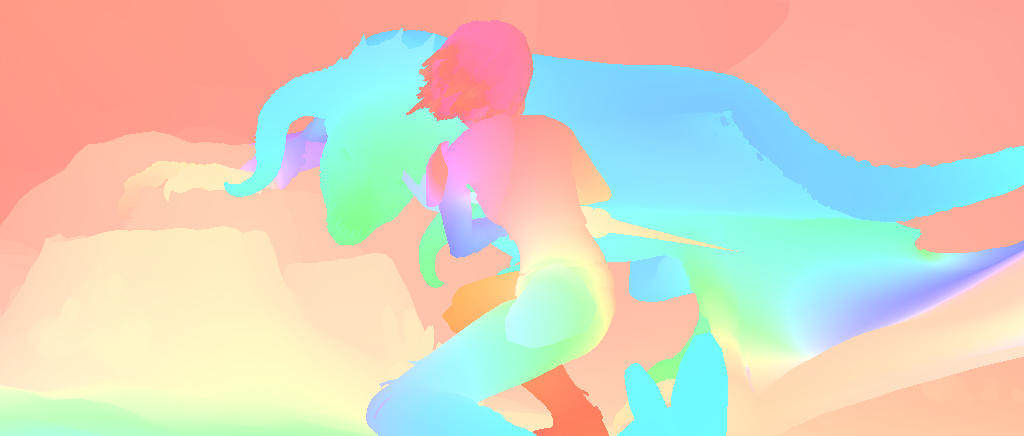} & 
			\includegraphics[width=0.19\linewidth]{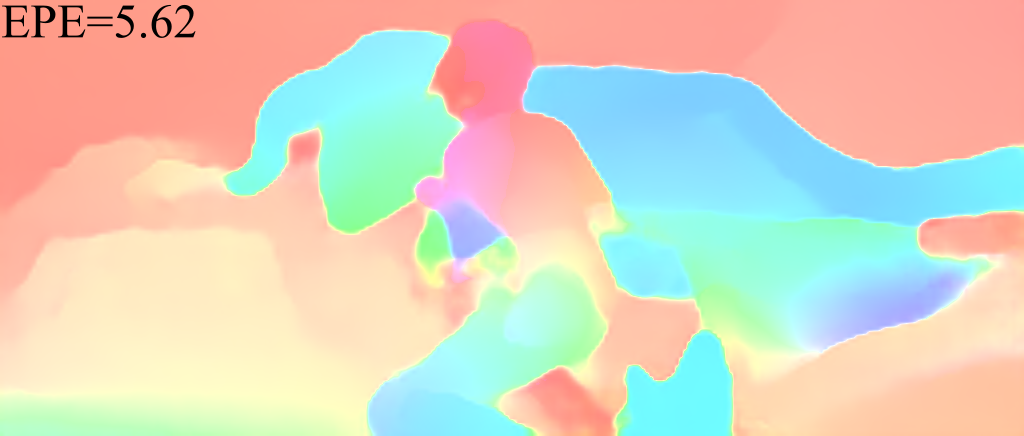} &
			\includegraphics[width=0.19\linewidth]{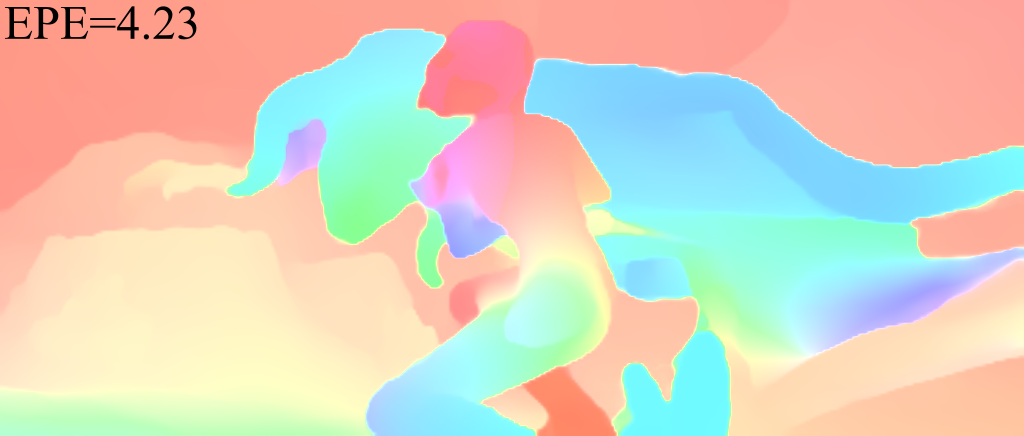} & 
			\includegraphics[width=0.19\linewidth]{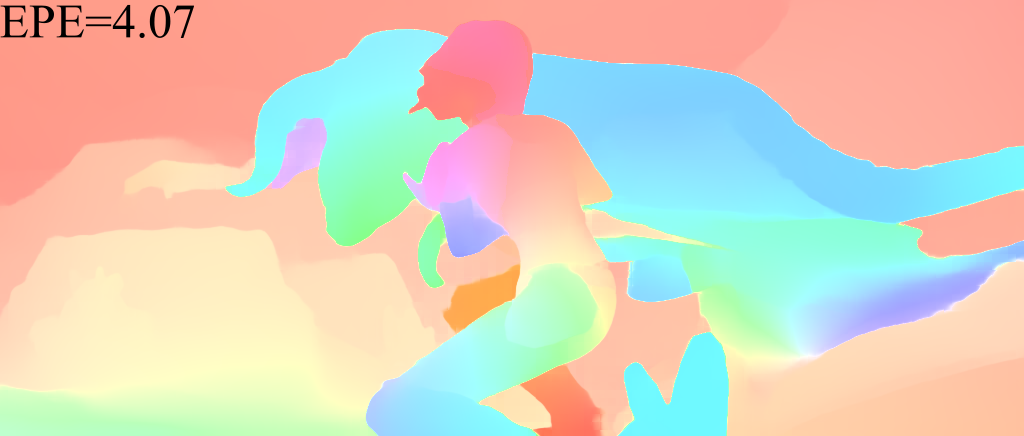} \\
			
			\includegraphics[width=0.19\linewidth]{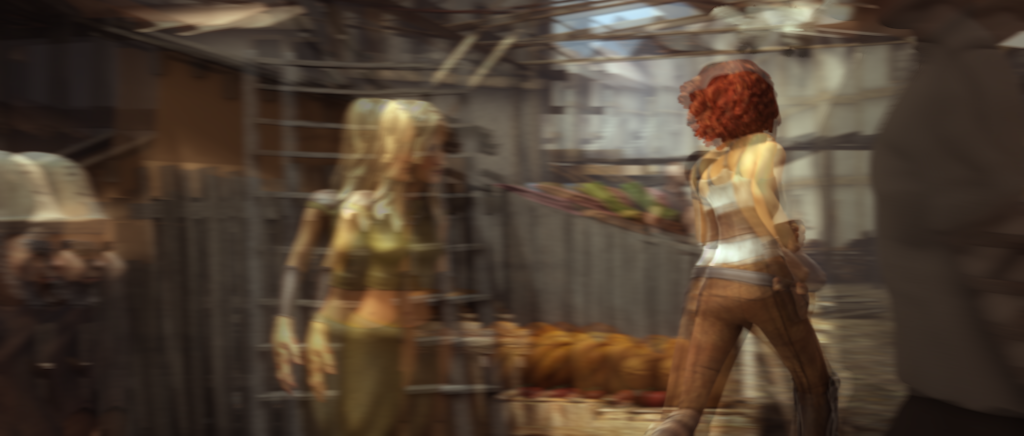} &
			\includegraphics[width=0.19\linewidth]{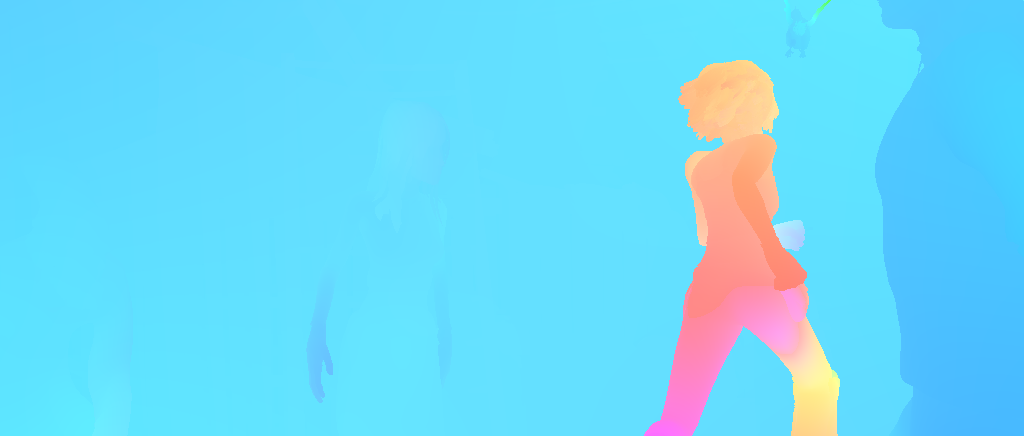} & 
			\includegraphics[width=0.19\linewidth]{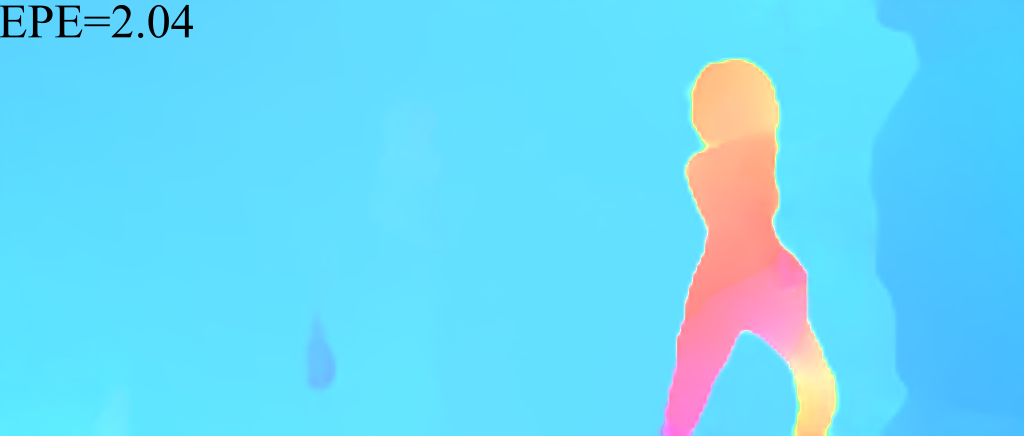} &
			\includegraphics[width=0.19\linewidth]{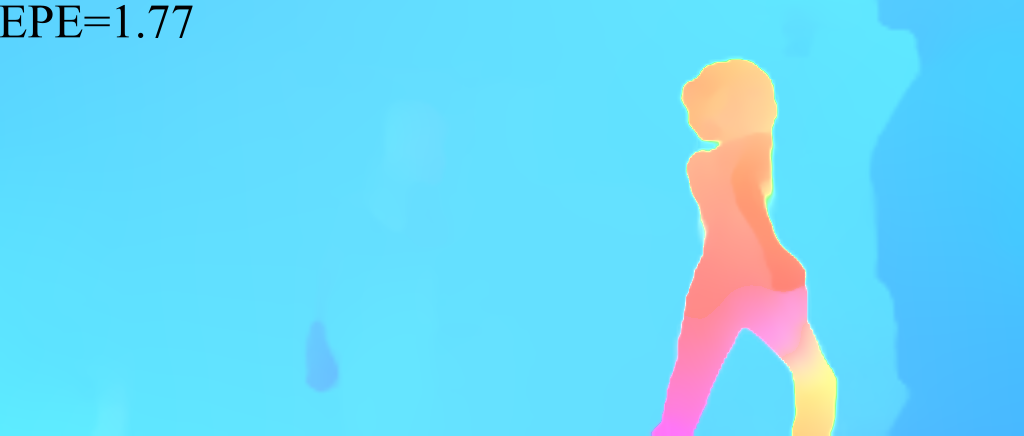} & 
			\includegraphics[width=0.19\linewidth]{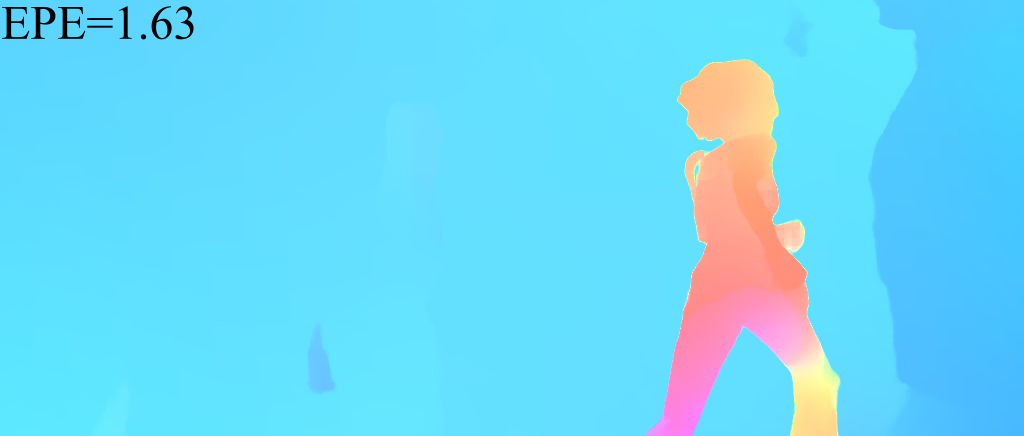} \\
			
			\includegraphics[width=0.19\linewidth]{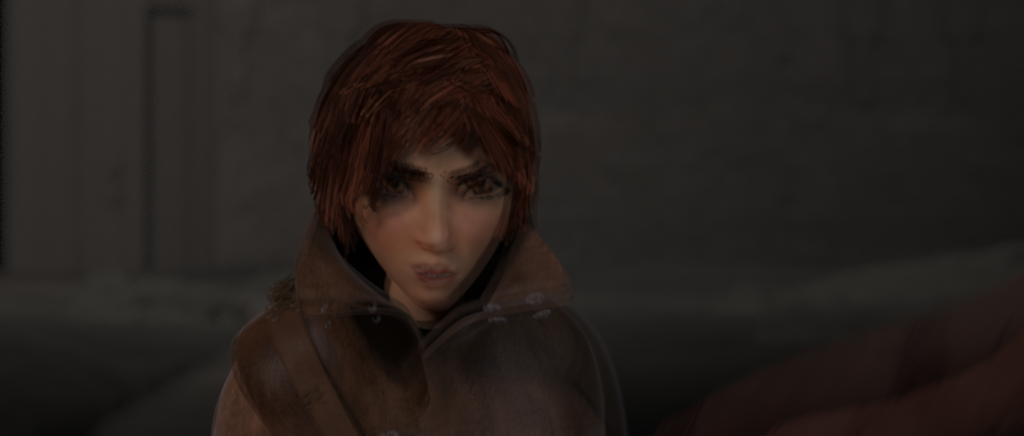} &
			\includegraphics[width=0.19\linewidth]{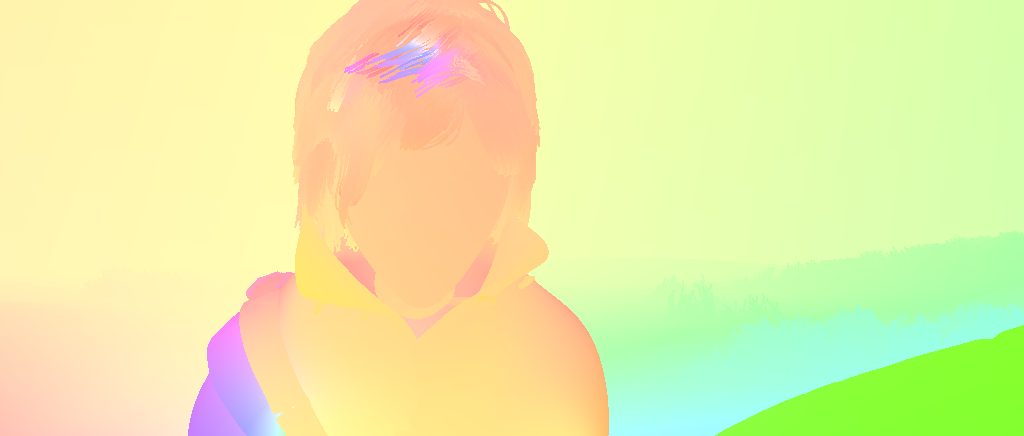} & 
			\includegraphics[width=0.19\linewidth]{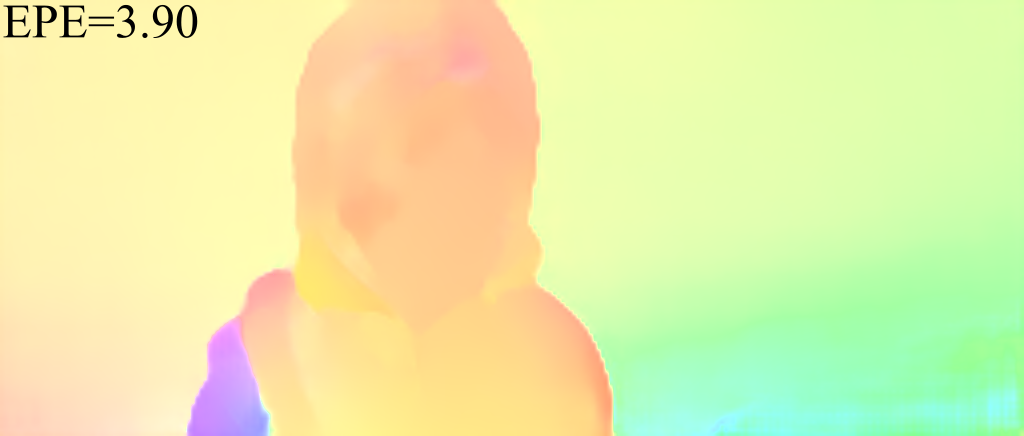} &
			\includegraphics[width=0.19\linewidth]{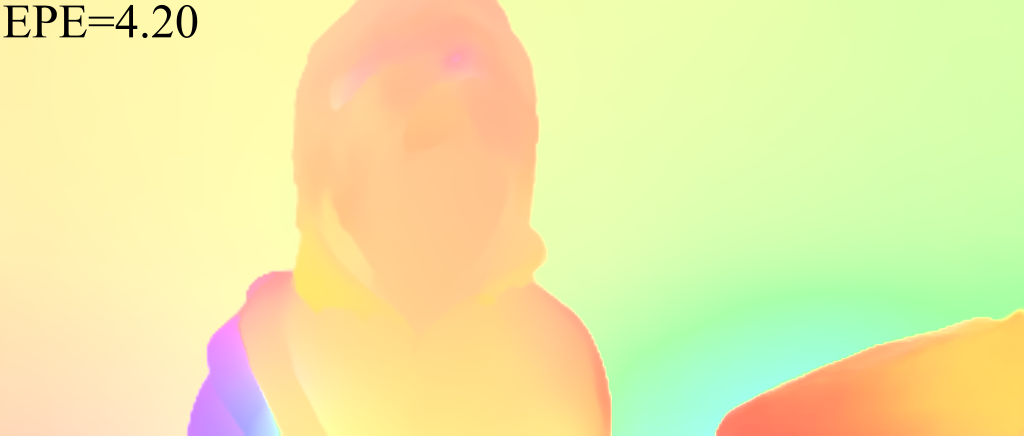} & 
			\includegraphics[width=0.19\linewidth]{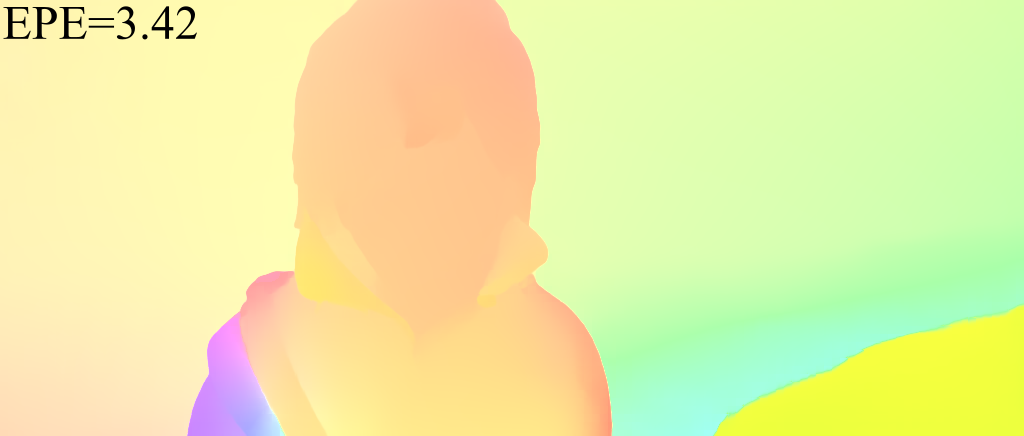} \\
			
			image overlay & ground truth & PWC-Net+ \cite{sun2019models} & MaskFlowNet \cite{zhao2020maskflownet} & Flow1D \\

		\end{tabular}
	}
	\caption{Visual results on Sintel test set.}
	\label{fig:vis_sintel_test}
\end{figure*}

We further show the visual comparison results with PWC-Net+ \cite{sun2019models} and MaskFlowNet \cite{zhao2020maskflownet} on Sintel test set in Fig.~\ref{fig:vis_sintel_test}.

\subsection{Implementation Details}

We use the same dataset schedule and hyper-parameters as RAFT \cite{teed2020raft} when training on FlyingChairs and FlyingThings3D datasets. For Sintel, we mix FlyingThings3D, KITTI 2015, HD1K and Sintel training set for additional fine-tuning. We random crop $368 \times 960$ resolutions as input and train for 100K iterations with a batch size of 6. For KITTI, we perform additional fine-tuning on KITTI 2015 training set for 50K iterations with a batch size of 6. The random crop size is $320 \times 1024$.

For training on very high-resolution images, we mix FlyingThings3D, Sintel, HD1K and Slow Flow \cite{janai2017slow} datasets for additional fine-tuning from Sintel weights. The Slow Flow dataset is created with high-speed camera and optimization is used to produce the `pseudo ground truth' flow. The resolutions of this dataset include $720 \times 1280$, $1024\times 1280$ and $576 \times 1024$. $3448$ image pairs in this dataset are used for training. To help our method generalize on 4K resolution images, we use larger crop size for training. Specifically, we random crop images in every mini-batch so that the resolutions are uniformly distributed between $640 \times 1080$ and $896 \times 1792$. For training images that are smaller than the crop size, we upsample them to the desired resolution, and the ground truth flow is upsampled accordingly. We train for 150K iterations with a batch size 2. All training is conducted on a single 32G V100 GPU.

\end{document}